\documentclass[runningheads]{llncs}

\usepackage{eccv}

\usepackage{eccvabbrv}

\usepackage{graphicx}
\usepackage{booktabs}

\usepackage{algorithm}
\usepackage{algpseudocode}   
\usepackage{setspace}

\usepackage{float}  

\usepackage{wrapfig}
\usepackage{amssymb}
\usepackage{pifont}
\usepackage{bbding}
\usepackage{makecell}

\usepackage{adjustbox}
\usepackage{multirow}
\usepackage{graphicx}
\usepackage{xfrac}

\usepackage{adjustbox}
\usepackage{subcaption}
\usepackage[table]{xcolor}
\usepackage{colortbl}
\usepackage{arydshln}
\makeatletter
\def\adl@drawiv#1#2#3{%
        \hskip.5\tabcolsep
        \xleaders#3{#2.5\@tempdimb #1{1}#2.5\@tempdimb}%
                #2\z@ plus1fil minus1fil\relax
        \hskip.5\tabcolsep}
\newcommand{\cdashlinelr}[1]{%
  \noalign{\vskip\aboverulesep
           \global\let\@dashdrawstore\adl@draw
           \global\let\adl@draw\adl@drawiv}
  \cdashline{#1}
  \noalign{\global\let\adl@draw\@dashdrawstore
           \vskip\belowrulesep}}
\makeatother

\newcommand\blfootnote[1]{%
  \begingroup
  \renewcommand\thefootnote{}\footnote{#1}%
  \addtocounter{footnote}{-1}%
  \endgroup
}

\usepackage{amssymb}
\usepackage{amsmath,amsfonts}
\usepackage{amsopn}
\usepackage{bm} %
\usepackage{multirow}
\usepackage{tabularx}

\newcommand{\Ours}{\method{CIVET}\xspace}

\newcommand{\ProbOpr}[1]{\mathbb{#1}}

\newcommand{\expect}[2]{%
\ifthenelse{\equal{#2}{}}{\ProbOpr{E}_{#1}}
{\ifthenelse{\equal{#1}{}}{\ProbOpr{E}\left[#2\right]}{\ProbOpr{E}_{#1}\left[#2\right]}}} %
\newcommand{\var}[2]{%
\ifthenelse{\equal{#2}{}}{\ProbOpr{VAR}_{#1}}
{\ifthenelse{\equal{#1}{}}{\ProbOpr{VAR}\left[#2\right]}{\ProbOpr{VAR}_{#1}\left[#2\right]}}} %

\newcommand{\method}[1]{\textsc{#1}}
\newcommand{\eat}[1]{}

\newcommand\mypara[1]{\vspace{.2cm}\noindent\textbf{#1\xspace}}

\definecolor{egogreen}{RGB}{214,235,197}
\definecolor{refred}{RGB}{251,210,204}
\definecolor{LightCyan}{rgb}{0.88,1,1}
\definecolor{LightRed}{rgb}{1,0.94,0.94}
\definecolor{LightBlue}{rgb}{0.94,0.97,1}
\definecolor{scarlet}{RGB}{147,0,0}
\definecolor{citeblue}{RGB}{238,26,28}
\definecolor{Blue}{RGB}{51,10,154}

\definecolor{citecolor}{RGB}{50, 63, 138}
\definecolor{linkcolor}{RGB}{187,18,26}
\usepackage[accsupp]{axessibility}  % Improves PDF readability for those with disabilities.

\usepackage[pagebackref,breaklinks,colorlinks,citecolor=eccvblue]{hyperref}
\usepackage{orcidlink}

\begin{document}

% \title{Author Guidelines for ECCV Submission} 

\title{\emph{When the City Teaches the Car:}\\Label-Free 3D Perception from Infrastructure}

\titlerunning{Label-Free 3D Perception from Infrastructure}

\author{Zhen Xu*\inst{1} \and
Jinsu Yoo*\inst{1} \and
Cristian Bautista* \inst{1} \and
Zanming Huang \inst{1} \and
Tai-Yu Pan \inst{2} \and
Zhenzhen Liu \inst{3} \and
Katie Z Luo \inst{4} \and
Mark Campbell \inst{3} \and
Bharath Hariharan \inst{3} \and
Wei-Lun Chao \inst{1,5}
}
\authorrunning{Z. Xu et al.}
\institute{
The Ohio State University \and
Google \and
Cornell University \and
Stanford University \and
Boston University
}
%******************
\maketitle
\blfootnote{$^*$Equal contribution. Project page: \url{https://jinsuyoo.info/civet}}

\begin{figure}
\centering
\includegraphics[width=\linewidth]{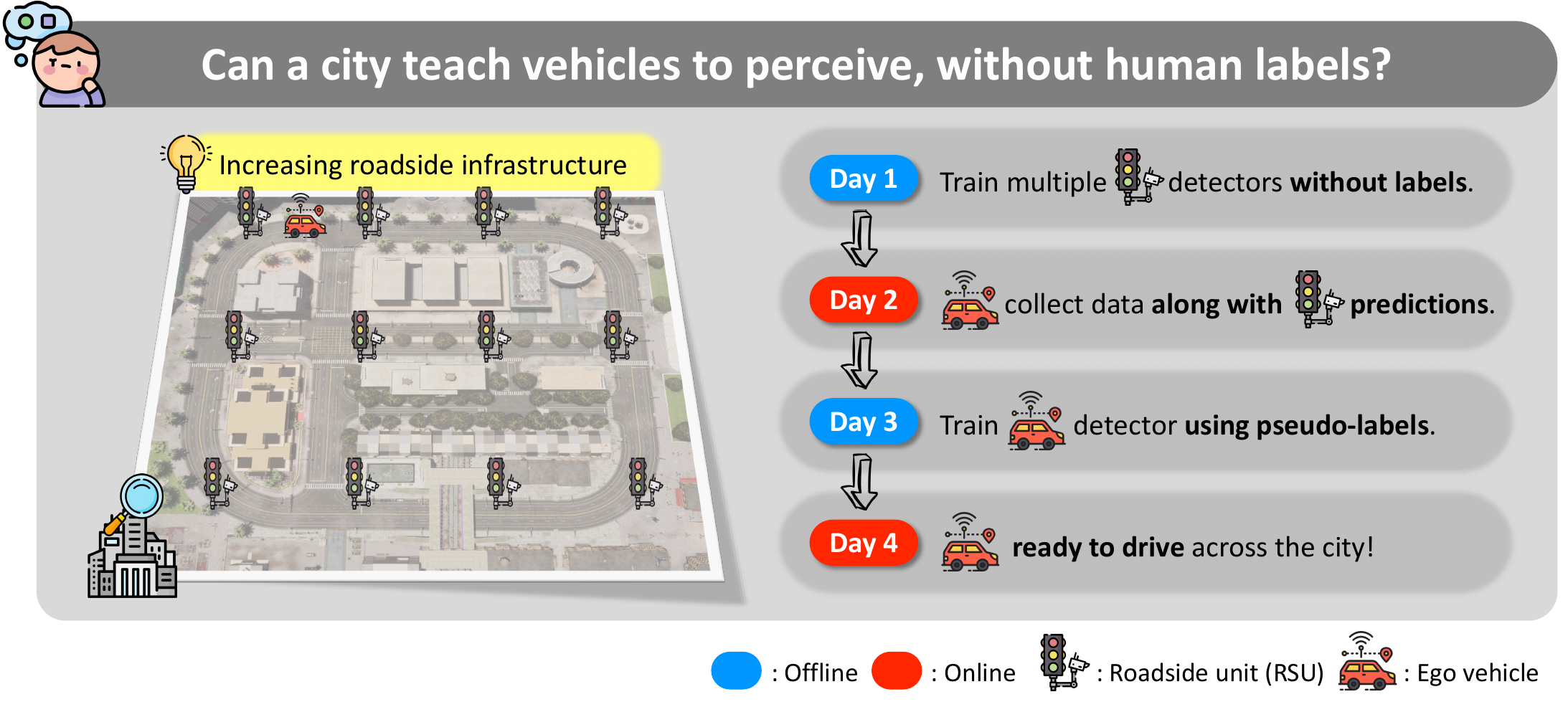}
% \vskip-15pt
\caption{
    \textbf{Can city infrastructure teach vehicles to perceive?} We explore a new paradigm where roadside infrastructure acts as distributed teachers, providing supervision to train ego perception models without manual annotations.
}
\vskip-25pt
\label{fig:teaser}
\end{figure}

\begin{abstract}
Building robust 3D perception for self-driving still relies heavily on large-scale data collection and manual annotation, yet this paradigm becomes impractical as deployment expands across diverse cities and regions. 
Meanwhile, modern cities are increasingly instrumented with \emph{roadside units} (RSUs), static sensors deployed along roads and at intersections to monitor traffic.
This raises a natural question: \emph{can the city itself help train the vehicle?}
We propose \emph{infrastructure-taught, label-free} 3D perception, a paradigm in which RSUs act as stationary, unsupervised teachers for ego vehicles. 
Leveraging their fixed viewpoints and repeated observations, RSUs learn local 3D detectors from unlabeled data and broadcast predictions to passing vehicles, which are aggregated as pseudo-label supervision for training a standalone ego detector. 
The resulting model requires no infrastructure or communication at test time.
We instantiate this idea as a fully \emph{label-free three-stage} pipeline and conduct a concept-and-feasibility study in a CARLA-based multi-agent environment. 
With CenterPoint, our pipeline achieves 82.3\% AP for detecting vehicles, compared to a fully supervised ego upper bound of 94.4\%.
We further systematically analyze each stage, evaluate its scalability, and demonstrate complementarity with existing ego-centric label-free methods. 
Together, these results suggest that city infrastructure itself can potentially provide a \emph{scalable supervisory signal} for autonomous vehicles, positioning infrastructure-taught learning as a promising \emph{orthogonal} paradigm for reducing annotation cost in 3D perception.
\end{abstract}

\section{Introduction}
\label{sec:intro}

Developing robust perception for self-driving is inherently challenging: the task is safety-critical, and driving scenes are highly diverse. Conventional practice scales ego-view data collection and annotation to improve generalization across deployment environments. However, traffic distributions vary substantially across cities and regions, reflecting differences in road layout, infrastructure, and traffic patterns. As deployment expands to new areas, repeated collect--label--retrain cycles are necessary, and simply scaling this process becomes infeasible for addressing the structural mismatch between finite labeled data and open-world complexity.

Meanwhile, modern cities are increasingly deploying roadside units (RSUs)---static sensors installed at intersections and along roads---to monitor traffic and support connected and cooperative intelligent transportation systems. 
For instance, transportation agencies in the United States and Europe have already begun deploying such infrastructure as part of large-scale connected-vehicle initiatives~\cite{usdot2023cvpilot,eu2023croads}. 
Unlike ego sensors, RSUs observe traffic from fixed vantage points with stable geometry and wide fields of view.

This raises a compelling possibility:
\emph{what if the city’s own sensing infrastructure (\eg, RSUs) could help train autonomous vehicles?}
More concretely, we propose turning increasingly prevalent RSUs into \emph{local perception experts} that provide supervision to passing ego vehicles. In principle, this could reduce the repeated burden of manual ego-side annotation and enable geographically localized supervision wherever such infrastructure is available. At a broader scale, one can envision a city-wide collection of RSUs serving as \emph{distributed annotators}, allowing vehicles to acquire supervision simply by driving through the city.

At first glance, this idea may seem overly ambitious: training a large number of RSUs into reliable perception systems could itself require prohibitively expensive labeled data. In that case, the burden of annotations would merely shift from the vehicle to the city, rather than being fundamentally reduced. Our key insight is that RSUs possess a property that makes this circle breakable: \emph{stationarity}. Because they repeatedly observe the same physical scene from fixed viewpoints over time, RSUs are more amenable to label-free learning based on temporal persistence, background stability, and recurring traffic patterns. Motivated by this insight, as shown in~\cref{fig:teaser}, we investigate infrastructure perception as a \emph{self-learning} teacher, leading to a fully label-free ecosystem for 3D perception. 

In this paper, we instantiate this new learning paradigm as \emph{infrastructure-taught, label-free} 3D perception and conduct a \emph{concept-and-feasibility} study. Because existing real-world multi-agent datasets remain limited in geographic diversity and coverage~\cite{luo2025mixedsignals,sekaran2025urbaning} and do not fully support controlled evaluation of our setting, we prepare a simulated dataset \Ours{} (\textbf{CI}ty-as-a-teacher for ego-\textbf{VE}hicle \textbf{T}raining), collected in a multi-agent environment built on CARLA~\cite{dosovitskiy2017carla} and V2Xverse~\cite{liu2025v2xverse}---following a common first step in V2X research, where simulation provides a reproducible testbed for principled study~\cite{xu2022opv2v,li2022v2xsim}. 
In a nutshell, \Ours{} spans four towns, each instrumented with 12 RSUs, covering both urban and rural environments (\cf \cref{sec:experiments_dataset}). 
Two towns provide contrasting traffic characteristics, while the remaining towns are incorporated to study the scalability of the proposed paradigm.

With this setup, we make the problem concrete and testable through a three-stage pipeline. 
In the first stage, each stationary RSU learns its own perception model from unlabeled observations by exploiting temporal consistency in dynamic traffic scenes. 
In the second stage, their predictions are broadcast to passing ego vehicles and aggregated as pseudo-label supervision for ego training. 
In the third stage, the ego detector is trained \emph{offline} and operates independently at test time, carrying distilled knowledge from the city.

We systematically evaluate the feasibility of this paradigm through baselines, bottleneck analysis, and upper-bound comparisons.
Within a town, our fully label-free pipeline achieves 82.3\% AP for detecting vehicles with CenterPoint~\cite{yin2021centerpoint}, compared to a fully supervised ego upper bound of 94.4\% (\cf~\cref{tab:stage3_ego}). 
This observation is consistent across different detectors~\cite{lang2019pointpillars,yin2021centerpoint} and geolocations. 
We further investigate scalability and geographic generalization by jointly training across four towns using all their RSUs (48 in total), and evaluating on the combined test set \emph{with greater diversity}, achieving 82.7\% AP for vehicles compared to an upper bound of 91.0\% (\cf~\cref{tab:general_ego}). 
Finally, we demonstrate complementarity with existing \emph{ego-centric} label-free methods. 
Combining pseudo-labels from a representative ego-centric approach (Oyster~\cite{zhang2023oyster}) with our infrastructure-generated supervision yields an additional $\sim$10 AP improvement over using either source alone (\cf~\cref{tab:ego_unsup_synergy}).
In summary, our contributions are three-fold:
\begin{itemize}
    \item We formulate a \textbf{new research direction}, \emph{infrastructure-taught 3D perception}, which investigates how city infrastructure can act as distributed teachers for training ego vehicles, potentially without manual supervision.
    
    \item We develop a \textbf{label-free pipeline} in which stationary RSUs learn local detectors from unlabeled observations and provide pseudo-label supervision to passing ego vehicles, producing a standalone ego detector at test time.
    
    \item We present the first \textbf{systematic concept-and-feasibility study} of this paradigm, including baselines, bottleneck analysis, upper bounds, and analyses of scalability and complementarity.
\end{itemize}

\mypara{Remark.}
Our setting is distinct from prior V2X and collaborative perception works~\cite{xu2023v2v4real,yu2022dair,li2024openmars,zimmer2024tumtraf,sekaran2025urbaning}, which perform online multi-agent fusion at inference time and rely on labeled training data. 
We use infrastructure only during training: RSUs act as unsupervised, stationary teachers whose predictions are distilled into a single ego-only model. 
Training is label-free, inference is ego-only, and no V2X link is needed at test time. 
Moreover, our goal is not to replace existing \emph{ego-centric} unsupervised approaches~\cite{you2022modest,zhang2023oyster,baur2024liso,wu2024cpd,lee2025openbox}, but to \emph{complement} them with a new supervision source. 
Rather than proposing another architecture within the standard ego-only regime, our paradigm changes \emph{where the training signals originate}, offering a complementary path toward more label-efficient learning.

\section{Related Work}
\label{sec:related}

\mypara{3D Object Detection.}
Accurate 3D detection is a fundamental building block for reliable autonomous driving. 
With the release of large-scale benchmarks~\cite{geiger2012kitti,caesar2020nuscenes,sun2020waymo}, researchers have developed a wide range of approaches, including LiDAR-based~\cite{zhou2018voxelnet,yan2018second,lang2019pointpillars,zhao2021point,yang2018ipod,yin2021centerpoint,shi2020pvrcnn,shi2019pointrcnn}, camera-based~\cite{li2024bevformer,li2023bevdepth,lin2022sparse4d,yang2023bevformer,wang2023streampetr,huang2022bevdet4d,liu2023sparsebev,li2023bevstereo}, and fusion-based~\cite{bai2022transfusion,liu2022bevfusion,xie2023sparsefusion,chen2017mv3d} methods.  
These models typically rely on large-scale labeled training data and often suffer from domain shifts when deployed in environments that differ from their training domains.
Our proposed paradigm facilitates scaling to new environments by providing supervision without manual annotation.

\mypara{Label-Efficient Learning.}
To alleviate the high cost of manual annotations---particularly for 3D point cloud data---numerous label-efficient methods have been explored. 
Representative directions include unsupervised learning~\cite{you2022hindsight,you2022modest,luo2023drift,najibi2022motion,zhang2023oyster,choy20194d,baur2024liso,seidenschwarz2024semoli,yang2021st3d,wu2024cpd,lentsch2024union,wu2025motal,zhang2025ua3d,xia2025learning,wu2025cpdplusplus}, semi-supervised learning~\cite{wang20213dioumatch,zhang2023simple,liu2022multimodal,xia2023coin,yang2024mixsup,xia2024hinted,liu2022ss3d,zhao2025sp3d}, and domain adaptation~\cite{wang2020germany,chen2024diffubox,yang2021st3d,yang2022st3d++}, as well as offboard detection~\cite{qi2021offboard,ma2023detzero}.  
Despite remarkable progress, these approaches remain fundamentally \emph{ego-centric}: supervision originates from the ego vehicle’s own observations. 
Recent work~\cite{yoo2025rnbpop} has shown that leveraging perception from a nearby expert vehicle can reduce ego annotation cost. We extend this idea to a city-wide, fully label-free ecosystem by treating distributed \emph{self-learning} RSUs as complementary teachers that collectively supervise the ego vehicle.

\mypara{Multi-Agent Collaborative Perception.}
Collaborative perception has gained attention as a means to mitigate occlusion and range limitations inherent to single-agent sensing.  
Typical approaches enable agents (\eg, vehicles and infrastructure) to exchange raw sensory data, intermediate neural features, or predicted objects to achieve better scene understanding~\cite{chen2019f,xu2022v2x,xu2022cobevt,xu2023v2v4real,yu2022dair,wang2020v2vnet,lu2024heal,hong2024multi,hu2024communication,pan2025typ,gao2025stamp,tang2025cost,zhou2025turbotrain,xu2025cosdh,yuan2025sparsealign,song2025trafalign}.  
In contrast, our work uses infrastructure during training rather than for inference-time fusion. The resulting ego model remains standalone at deployment, without requiring communication at test time.

\mypara{Datasets for Collaborative Driving.}
Recent progress has been driven largely by community efforts to collect V2X datasets---covering vehicle-to-vehicle~\cite{xu2022opv2v,xu2023v2v4real,li2024openmars}, vehicle-to-infrastructure~\cite{li2022v2xsim,xu2022v2x,yu2022dair,ma2024holovic,xiang2024v2xreal,zimmer2024tumtraf,luo2025mixedsignals,liu2025v2xverse,sekaran2025urbaning}, and infrastructure-to-infrastructure~\cite{hao2024rcooper} settings.  
However, even recent real-world datasets remain geographically limited---often covering only a handful of intersections and RSUs~\cite{luo2025mixedsignals,sekaran2025urbaning}---due to hardware deployment cost, maintenance overhead, and regulatory constraints.  
Such limitations make it difficult to systematically study our proposed paradigm, in which a city-wide collection of RSUs can serve as teachers for label-free ego training. 
Following the common simulation-to-real trajectory in V2X research, we begin by studying the proposed paradigm in a controlled simulation environment to assess its feasibility.

\begin{figure*}[t]
    \centering
    \includegraphics[width=\linewidth]{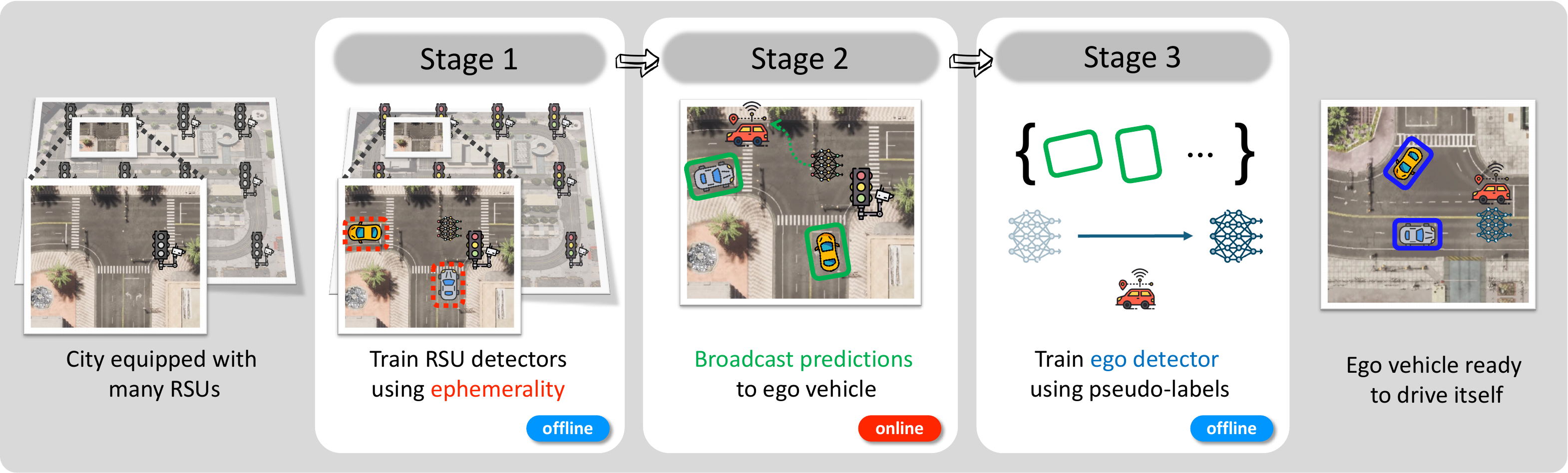}
\caption{\textbf{Overview of infrastructure-taught, label-free 3D perception.}
Stage~1: each RSU learns a location-specialized detector in an unsupervised manner by exploiting temporal consistency from its stationary viewpoint. 
Stage~2: trained RSUs broadcast their predicted 3D bounding boxes to nearby ego vehicles when their fields of view overlap. 
Stage~3: the ego vehicle aggregates these predictions as pseudo-labels to train its own detector offline, producing a standalone ego model that no longer requires infrastructure at deployment time.}
    %\vskip-5pt
    \label{fig:pipeline}
\end{figure*}

\section{Infrastructure-Taught, Label-Free 3D Perception}

We instantiate \emph{infrastructure-taught, label-free} 3D perception as a fully label-free framework for training an ego 3D detector using roadside infrastructure as supervision. At a high level, the framework operates in three stages (see~\cref{fig:pipeline}): (1) each RSU learns a detector from its own unlabeled observations, (2) trained RSUs broadcast predictions to passing ego vehicles, and (3) the ego detector is trained using these infrastructure-generated pseudo-labels.

\subsection{Problem Setup}

We consider a geo-fenced city instrumented with a set of RSUs, denoted as $\mathcal{R} = \{ R_1, R_2, \ldots, R_M \}$.
Each $R_i$ is equipped with sensors (\eg, cameras and LiDAR) and a local 3D detector $f_R^i$. Leveraging their static viewpoints, RSUs train these detectors in a label-free manner using temporal consistency. The resulting detectors produce predictions $Y_R^i = f_R^i(X_R^i)$ from sensory inputs $X_R^i$.

An ego vehicle $E$ drives through the same city, collecting its own sensory data $X_E$. Whenever it is within the communication range of an RSU, it receives predictions from nearby RSUs, which are transformed into ego-centric coordinates using the known rigid transformation $\mathcal{T}_{R_i \rightarrow E}$. The ego aggregates valid predictions from nearby RSUs---after filtering and fusing overlapping boxes---to obtain pseudo-labels $\mathcal{Y}_E$. Together with $X_E$, these form the training set $\mathcal{D}_E = \{ (X_E, \mathcal{Y}_E) \}$ used to train the ego’s own detector $f_E$ offline.

Our goal is to train an ego-centric detector $f_E$ using only label-free supervision transferred from the RSUs, such that it can operate independently at deployment without manual labels or infrastructure communication.

\subsection{Overall Pipeline}

\mypara{Stage 1: Unsupervised RSU Training.}
\label{sec:rsu_training}
Manual annotation may be feasible for a handful of RSUs, but it becomes impractical as infrastructure scales. Thus, the first stage of our pipeline enables each RSU to train its own detector $f_R^i$ in a fully label-free manner, using only its continuous local observations.

We exploit the \emph{static nature of RSUs}: their viewpoints remain constant over time, allowing them to repeatedly observe the same environment under varying traffic dynamics. In such settings, background regions remain temporally consistent, while dynamic objects---such as vehicles, cyclists, and pedestrians---appear and disappear across frames. This temporal contrast provides a natural self-supervised cue for discovering moving objects without human labels.

Concretely, for each RSU $R_i \in \mathcal{R}$ equipped with a LiDAR sensor, we adopt the \textit{persistence point score} (PP score) proposed in MODEST~\cite{you2022modest} to quantify the temporal stability of points across repeated observations of the same scene. Points with low persistence are identified as likely dynamic and are used as seeds for object proposals. We apply DBSCAN~\cite{ester1996dbscan} to cluster these transient points into coarse pseudo-labels for each frame. 

To further improve the spatial and temporal consistency of the generated boxes, we incorporate a tracking method~\cite{zhang2023oyster} that smooths trajectories and suppresses spurious initial boxes. Specifically, we track each object and select its most reliable annotation---the frame in which the object is closest to the RSU in the $x$--$y$ plane. The box size from this frame is then used to adjust the sizes of all remaining boxes corresponding to the same object. The resulting pseudo-labels $Y_R^i$ serve as supervision to train each RSU detector $f_R^i$.

Through this process, each RSU becomes specialized to its own fixed field of view and produces local detections that later serve as supervision for the ego detector. As shown in~\cref{fig:exp_cross_eval_rsu}, these unsupervised RSU detectors perform strongly within their own views, achieving 82.8\% AP on average across all RSUs. This local specialization forms the basis for the next RSU-to-ego supervision stage.

\begin{algorithm}[t]
\small
\setstretch{1.1}
\caption{Aggregating Knowledge from Many RSUs}
\label{alg:stage2}
\begin{algorithmic}[1]
    \Require Trained RSU detectors $\{f_R^i\}_{i=1}^M$, ego trajectory frames $\{t=1..T\}$, maximum RSU distance $d_{\max}$
    \State Initialize ego pseudo-label dataset $\mathcal{D}_E = \emptyset$
    \For{each time step $t = 1$ to $T$}
        \State Acquire ego sensory data $X_E(t)$
        \State Initialize temporary collection $\mathcal{C} = \emptyset$
        \For{each RSU $R_i$ such that $\text{dist}(E_t, R_i) < d_{\max}$}
            \State Obtain RSU prediction $Y_R^i = f_R^i(X_R^i(t))$
            \State Transform to ego frame $\tilde{Y}_R^i = \mathcal{T}_{R_i \rightarrow E}(Y_R^i)$
            \State $\mathcal{C} \gets \mathcal{C} \cup \tilde{Y}_R^i$
        \EndFor
        \State Aggregate bounding boxes $\mathcal{C}$ to obtain $\mathcal{Y}_E(t)$
        \State Store $(X_E(t), \mathcal{Y}_E(t))$ into $\mathcal{D}_E$
    \EndFor
    \State \Return $\mathcal{D}_E$
\end{algorithmic}
\end{algorithm}

\mypara{Stage 2: Label Transfer and Aggregation.}
\label{sec:broadcasting}
After the RSUs have been trained, the ego vehicle traverses the town to collect sensory data for training its own detector. Instead of relying on manual annotations, we transfer supervision from infrastructure: each RSU, equipped with its self-learned detector, broadcasts its predictions to the ego vehicle in an online manner (see \cref{alg:stage2}).

Specifically, when the ego enters the field of view of an RSU $R_i \in \mathcal{R}$, the RSU transmits its predictions $Y_R^i = f_R^i(X_R^i)$, which are transformed into the ego’s coordinate using $\mathcal{T}_{R_i \rightarrow E}$. The ego aggregates all such predictions as pseudo-labels aligned with its own sensory data, thereby collecting supervision as it drives.

To improve pseudo-label quality, boxes that are too distant or sparsely populated by LiDAR points (\eg, due to occlusion) are filtered out.\footnote{We observe that removing filtering drops AP from 66.4 in~\cref{tab:stage3_ego}~\ding{185} to 60.0.} In regions with overlapping RSU coverage, multiple predictions for the same object may occur. We address this with distance-weighted non-maximum suppression (NMS), which scores each box by the inverse distance between the RSU and the detected object, thereby prioritizing closer and typically more reliable observations. This yields a unified pseudo-label set $\mathcal{Y}_E$. We note that more sophisticated aggregation strategies are possible; however, we keep the aggregation module simple yet workable, as our goal is to study the feasibility of infrastructure-taught supervision rather than to optimize this component.

By repeating this process, the ego vehicle constructs its own pseudo-labeled dataset $\mathcal{D}_E$ without any manual annotation. As shown in~\cref{tab:stage2_broadcast}, the RSUs collaboratively provide high-quality pseudo-labels for the ego vehicle, achieving recall of 89.5\%, 73.7\%, and 94.7\% for cars, pedestrians, and cyclists, respectively.

\mypara{Stage 3: Ego Detector Training.}
\label{sec:ego_training}
With pseudo-labels $\mathcal{D}_E = \{(X_E, \mathcal{Y}_E)\}$ gathered across the city, the ego vehicle trains its own 3D detector $f_E$ entirely offline. Unlike the RSU specialists, which are tied to fixed viewpoints, the ego detector integrates supervision transferred from multiple regions into a single model. Once trained, it operates independently at test time without requiring RSUs or manual annotations.

\mypara{Putting It Together.}
Together, these three stages define a framework for infrastructure-taught, label-free 3D perception: RSUs first learn from their own unlabeled observations, their predictions are then transferred to the ego as pseudo-labels, and the ego detector is finally trained as a standalone model.

It is worth noting that we focus on evidence of feasibility rather than algorithmic advancement of individual modules: each module is intentionally a simple-yet-workable choice. We believe advances in any module (from pseudo-labeling to broadcasting to supervised training) will directly benefit our pipeline.

\subsection{\Ours Dataset}

\begin{figure*}[t]
    \centering
    \includegraphics[width=\linewidth]{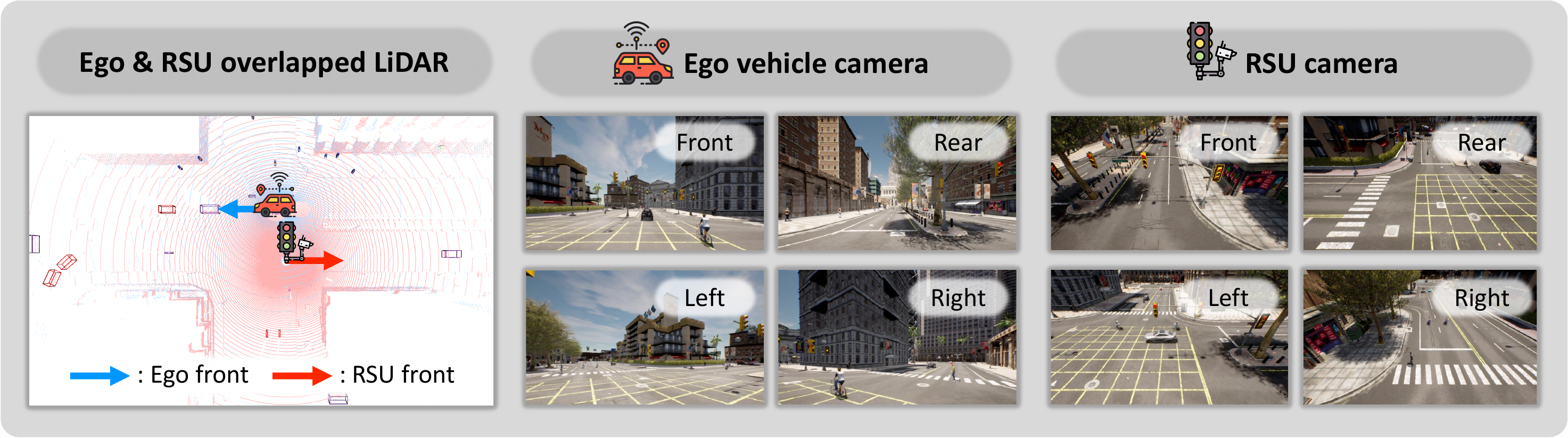}
    \caption{\textbf{Sample from the \textsc{CIVET} dataset used in Stage~2 RSU-to-ego broadcasting.} The ego vehicle and RSU observe the same traffic scene from different viewpoints with overlapping fields of view.}
    \label{fig:sample_data}
    % \vskip-5pt
\end{figure*}

To study the proposed paradigm under controlled conditions, we construct a simulated multi-agent dataset, \Ours{}, built on CARLA~\cite{dosovitskiy2017carla} using V2Xverse~\cite{liu2025v2xverse}.
Our goal is not merely to collect multi-agent data, but to create a controlled setting that enables systematic analysis under varying RSU configurations and geographic layouts.
In particular, we require an environment that (1) enables controlled manipulation of hardware, such as RSU placement, and (2) allows us to study how multiple RSUs can jointly supervise an ego vehicle under different scene distributions. \Ours{} is designed to support these analyses across four geo-fenced towns instrumented with many RSUs. Two towns provide contrasting traffic characteristics (\eg, different object scales, densities, and layouts) used in our core experiments, while the remaining towns are included to study scalability of the proposed paradigm.
An example scene is shown in~\cref{fig:sample_data}. Both RSUs and the ego vehicle observe the environment using cameras and LiDAR. We prepare separate data splits for RSU training, RSU-to-ego broadcasting, and ego-detector evaluation; details are provided in \cref{sec:experiments_dataset} and \textcolor{red}{Suppl. Sec. B}.

\section{Experiments}

\subsection{Data Collection}
\label{sec:experiments_dataset}

\mypara{Town Selection.}
Data are collected in CARLA~\cite{dosovitskiy2017carla} Town~1,~2,~7, and~10. 
Among them, Town~7 and~10 provide complementary rural and urban environments used in our core experiments. 
To better reflect diversity in deployment conditions, these two towns differ not only in layout but also in object scale and traffic composition: Town~10 contains relatively smaller cars and denser pedestrian and cyclist traffic, whereas Town~7 features larger vehicle models and sparser human activity. 
Town~1 and~2 are additionally included to study scalability.

\mypara{Sensor Configuration.}
We deploy 12 RSUs in each town at diverse locations to enable multi-RSU supervision under varying spatial relationships with the ego vehicle. Each RSU is equipped with four RGB cameras, four depth cameras, and a 128-beam LiDAR. 
The simulator provides 3D bounding box annotations for three object categories: \emph{Car}, \emph{Pedestrian}, and \emph{Cyclist}, with consistent object IDs across frames.
The ego vehicle uses a similar sensor suite---four RGB cameras, four depth cameras, and a 64-beam LiDAR---and additionally has access to a top-down map view and the same box annotations.

\mypara{Dataset Size.}
For RSU training, we collect 7k frames per RSU (5k for training and 2k for validation), resulting in 84k RSU frames \emph{per town} across 12 RSUs. 
For RSU-to-ego broadcasting (ego training), we collect an additional 5k synchronized frames \emph{per town} in which the ego vehicle and multiple RSUs share overlapping fields of view, enabling pseudo-label transfer. 
A separate set of 3k ego frames \emph{per town} is reserved for evaluating the ego detector.
In total, the dataset contains 608k frames across four towns.

Since the data are collected in simulation, larger-scale datasets can be generated if needed. 
In practice, the current scale already enables systematic evaluation of the proposed paradigm and its components. 
We therefore keep the dataset size fixed in this study, while noting that scaling up data collection is straightforward in our reproducible simulation environment.

\subsection{Experimental Setups}
\label{sec:exp_setups}

We primarily report results on Town~10 (urban), which exhibits higher object density; additional results are provided in the \textcolor{red}{Suppl. Sec. D}.

\mypara{Implementation.}
All experiments are conducted using LiDAR-based 3D detectors. For unsupervised RSU training, we divide 5k training frames into 10 temporally equidistant segments and compute the PP score for each segment. Following~\cite{you2022modest}, we remove estimated ground points before obtaining initial pseudo-labeled bounding boxes. Each RSU detector adopts CenterPoint~\cite{yin2021centerpoint} and is trained for 20 epochs using an initial learning rate of $5\times10^{-4}$, which is reduced by a factor of 0.1 at epochs 10 and 15.

During RSU-to-ego broadcasting, we set the maximum communication range to 160\,m. Following~\cite{xu2022opv2v}, we simulate realistic communication noise by adding Gaussian perturbations to transmitted 3D positions ($\sigma = 0.2\,\mathrm{m}$) and yaw angles ($\sigma = 0.05\,\mathrm{rad}$), together with a fixed 100\,ms communication delay.\footnote{Communication imperfections are \emph{shared challenges} across V2X systems rather than being specific to our formulation. We therefore simulate representative noise conditions and explore a simple refinement module (\cf.~\cref{tab:stage3_ego}~\ding{176}\ding{186}). Advances in synchronization and calibration would directly benefit this setting.} To fuse overlapping boxes from multiple RSUs, we apply distance-weighted NMS with an IoU threshold of 0.1. Since each RSU detector outputs class-agnostic boxes, we further assign semantic categories using a template-based class matching scheme that compares box sizes to known object prototypes, inspired by~\cite{wu2024cpd}. For ego training, we use two 3D detectors, PointPillars~\cite{lang2019pointpillars} and CenterPoint~\cite{yin2021centerpoint}, trained with the same learning schedule and number of epochs as the RSU detectors. All experiments are conducted on two NVIDIA A100 GPUs.

\begin{table*}[t]
\centering
\caption{
\textbf{Systematic study of infrastructure-taught ego training.}
We vary RSU supervision and communication noise to analyze their impact on downstream ego detector performance.
``Refine'' denotes applying heuristic box refinement~\cite{luo2023drift} to improve pseudo-label quality before ego training.
}
\label{tab:stage3_ego}
% \vskip-10pt
\begin{adjustbox}{width=\linewidth,center}
\begin{tabular}{c l l c c c c c c}
\toprule
 & \multicolumn{2}{c}{Stage 1 (RSU training)} & \multicolumn{2}{c}{Stage 2 (Broadcasting)} & \multicolumn{4}{c}{Stage 3 (Ego testing)} \\
\cmidrule(lr){2-3} \cmidrule(lr){4-5} \cmidrule(lr){6-9}
~~ID~~ & Annotation source~~~ & Method~~~ & ~~Comm.\ noise~~ & ~~Refine~~ & Car & Ped. & Cyc. & \cellcolor{blue!10}Avg. \\
\midrule

\rowcolor{gray!10} \multicolumn{9}{l}{~~\emph{Ego detector: PointPillars~\cite{lang2019pointpillars}}}\\

\ding{172} & 12 RSUs (unsup.) & PP score & -- & -- & 74.3 & 77.0 & 68.0 & \cellcolor{blue!10}73.1 \\
\ding{173} & 12 RSUs (unsup.) & PP score & $\checkmark$ & -- & 77.1 & 49.4 & 57.0 & \cellcolor{blue!10}61.2 \\

\cmidrule{2-9}

\ding{174} & 12 RSUs (unsup.) & PP score + tracking & -- & -- & 79.3 & 77.5 & 79.0 & \cellcolor{blue!10}78.6 \\
\ding{175} & 12 RSUs (unsup.) & PP score + tracking & $\checkmark$ & -- & 80.3 & 49.1 & 69.6 & \cellcolor{blue!10}66.3 \\

\cmidrule{2-9}

\ding{176} & 12 RSUs (unsup.) & PP score + tracking & $\checkmark$ & $\checkmark$ & 76.5 & 52.7 & 79.8 & \cellcolor{blue!10}69.7 \\

\cmidrule{2-9}

\ding{177} & \textcolor{gray}{12 RSUs (sup.)} & \textcolor{gray}{--} & \textcolor{gray}{--} & \textcolor{gray}{--} & \textcolor{gray}{92.1} & \textcolor{gray}{82.2} & \textcolor{gray}{92.8} & \cellcolor{blue!10}\textcolor{gray}{89.0} \\
\ding{178} & \textcolor{gray}{12 RSUs (sup.)} & \textcolor{gray}{--} & \textcolor{gray}{$\checkmark$} & \textcolor{gray}{--} & \textcolor{gray}{90.1} & \textcolor{gray}{40.3} & \textcolor{gray}{86.2} & \cellcolor{blue!10}\textcolor{gray}{72.2} \\

\cmidrule{2-9}

$\star$ & \textcolor{gray}{Ego ground-truth} & \textcolor{gray}{--} & \textcolor{gray}{--} & -- & \textcolor{gray}{94.4} & \textcolor{gray}{88.0} & \textcolor{gray}{93.7} & \cellcolor{blue!10}\textcolor{gray}{92.0} \\

\midrule

\rowcolor{gray!10} \multicolumn{9}{l}{~~\emph{Ego detector: CenterPoint~\cite{yin2021centerpoint}}} \\

\ding{182} & 12 RSUs (unsup.) & PP score & -- & -- & 78.7 & 78.3 & 61.1 & \cellcolor{blue!10}72.7 \\
\ding{183} & 12 RSUs (unsup.) & PP score & $\checkmark$ & -- & 79.2 & 45.2 & 49.5 & \cellcolor{blue!10}58.0 \\

\cmidrule{2-9}

\ding{184} & 12 RSUs (unsup.) & PP score + tracking & -- & -- & 82.3 & 79.3 & 68.5 & \cellcolor{blue!10}76.7 \\
\ding{185} & 12 RSUs (unsup.) & PP score + tracking & $\checkmark$ & -- & 80.4 & 52.3 & 66.4 & \cellcolor{blue!10}66.4 \\

\cmidrule{2-9}

\ding{186} & 12 RSUs (unsup.) & PP score + tracking & $\checkmark$ & $\checkmark$ & 77.7 & 56.0 & 75.7 & \cellcolor{blue!10}69.8 \\

\cmidrule{2-9}

\ding{187} & \textcolor{gray}{12 RSUs (sup.)} & \textcolor{gray}{--} & \textcolor{gray}{--} & -- & \textcolor{gray}{93.9} & \textcolor{gray}{84.4} & \textcolor{gray}{93.4} & \cellcolor{blue!10}\textcolor{gray}{90.6} \\
\ding{188} & \textcolor{gray}{12 RSUs (sup.)} & \textcolor{gray}{--} & \textcolor{gray}{$\checkmark$} & -- & \textcolor{gray}{92.4} & \textcolor{gray}{35.9} & \textcolor{gray}{87.3} & \cellcolor{blue!10}\textcolor{gray}{71.9} \\

\cmidrule{2-9}

$\star$ & \textcolor{gray}{Ego ground-truth} & \textcolor{gray}{--} & \textcolor{gray}{--} & -- & \textcolor{gray}{94.4} & \textcolor{gray}{91.8} & \textcolor{gray}{96.6} & \cellcolor{blue!10}\textcolor{gray}{94.3} \\

\bottomrule
\end{tabular}
\end{adjustbox}
\end{table*}

\mypara{Evaluation.}
For evaluation of the unsupervised RSU detectors, we follow MODEST~\cite{you2022modest} and merge the \textit{Car}, \textit{Pedestrian}, and \textit{Cyclist} categories into a single foreground class. For ego detector evaluation, unless otherwise specified, we report bird’s-eye-view (BEV) average precision (AP) at IoU thresholds of 0.5 for cars and 0.3 for pedestrians and cyclists. The detection region is defined as $[-80, 80]$\,m longitudinally for both RSU and ego detectors, while the lateral range is $[-80, 80]$\,m for RSUs and $[-40, 40]$\,m for the ego detector.

\subsection{Main Results and Analysis \small{(Qualitative Results in \textcolor{red}{Suppl. Sec. D.3})}}

We evaluate multiple configurations of the proposed pipeline to understand the key factors governing infrastructure-taught ego learning. 
The main quantitative results are reported in~\cref{tab:stage3_ego}, and we elaborate key findings below.

First, comparing \ding{174}\ding{184} with the ego upper bound ($\star$), we find that the fully label-free pipeline is \emph{feasible} in practice, achieving competitive performance across object categories for both PointPillars~\cite{lang2019pointpillars} and CenterPoint~\cite{yin2021centerpoint}. This indicates that infrastructure-taught supervision can recover a substantial portion of the supervised upper bound.\footnote{Interestingly, in some cases the unsupervised RSU setup outperforms its supervised counterpart for pedestrians. We attribute this to unsupervised RSUs generating a larger number of pseudo-labels, which may provide stronger supervision for the ego.}

Second, comparing \ding{172}\ding{182}, \ding{174}\ding{184}, and \ding{177}\ding{187}, we find that pseudo-labels generated by unsupervised RSU detectors provide effective supervision for ego training, approaching the performance of pseudo-labels produced by \emph{supervised RSU detectors}. Incorporating tracking consistently improves ego performance.

Third, comparing \ding{172}\ding{182} with \ding{173}\ding{183}, and \ding{174}\ding{184} with \ding{175}\ding{185}, we observe that simulated communication imperfections degrade ego detector performance, with the largest drop occurring for pedestrians. Since pedestrians are smaller objects, they are more sensitive to RSU-to-ego misalignment, which is also reflected in the pseudo-label quality reported in~\cref{tab:stage2_broadcast}. This highlights pseudo-label refinement as an important direction for future work.

Finally, we explore a baseline strategy to mitigate communication noise. 
Inspired by~\cite{luo2023drift}, we apply a heuristic box refinement method that samples candidate boxes around each pseudo-label and selects the highest-scoring candidate using heuristics (\eg, the number of points inside the box and boundary alignment; details in \textcolor{red}{Suppl. Sec.~C.2}). Comparing \ding{175}\ding{185} with \ding{176}\ding{186}, we observe consistent average AP improvements. This suggests that improved alignment and calibration techniques from V2X could further benefit our infra-taught ego learning.

\begin{figure*}[t]
    \centering
    \begin{minipage}[t]{0.48\linewidth}
        \centering
        \includegraphics[width=\linewidth]{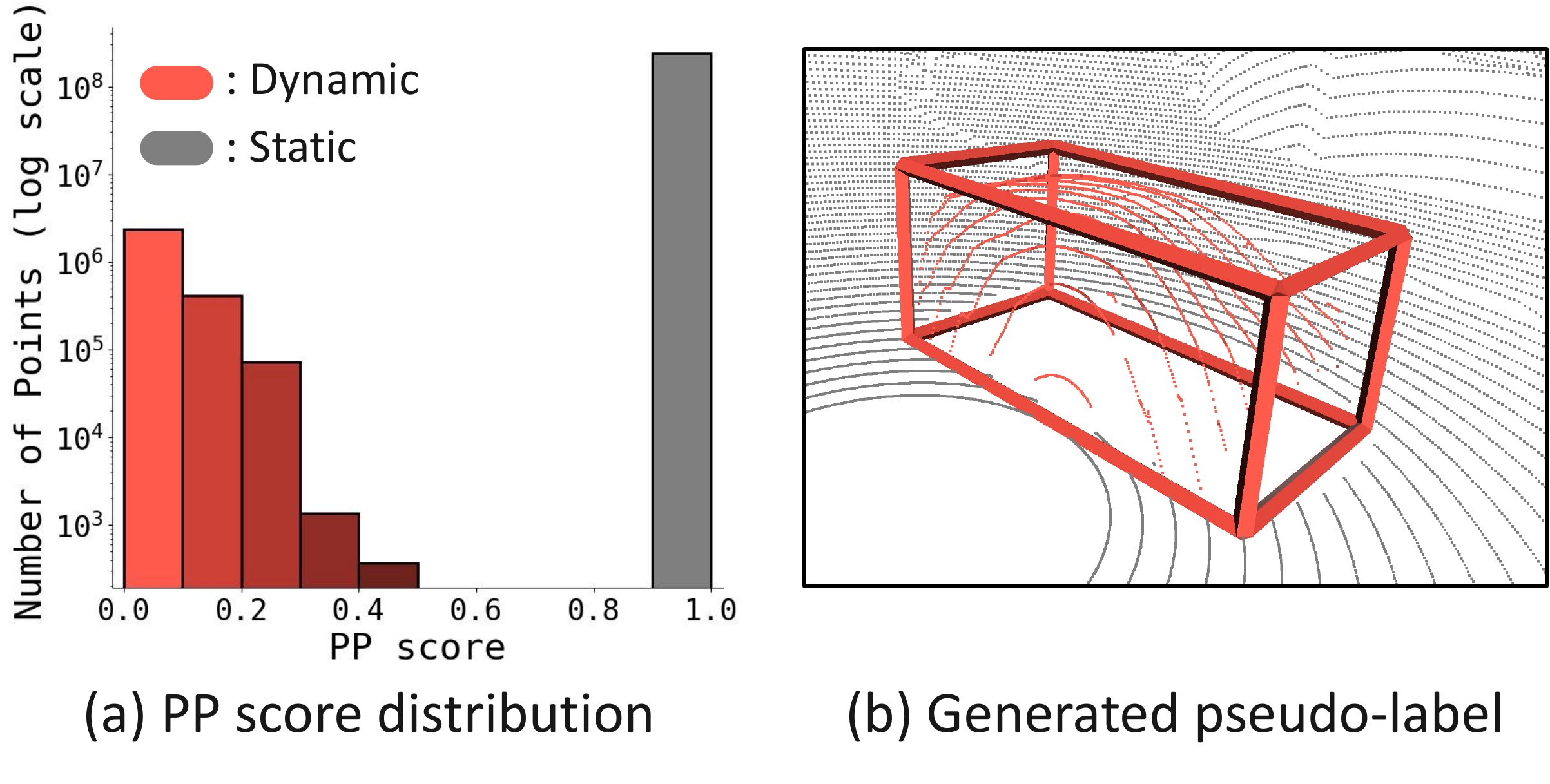}
        \vskip-5pt
        \caption{\textbf{Effectiveness of PP scores for RSU.}
        (a) Discriminative distribution allows a clear separation between static background and objects.
        (b) Pseudo-labels exhibit high localization quality.}
        \label{fig:pp_score}
    \end{minipage}\hfill
    \begin{minipage}[t]{0.48\linewidth}
        \centering
        \includegraphics[width=\linewidth]{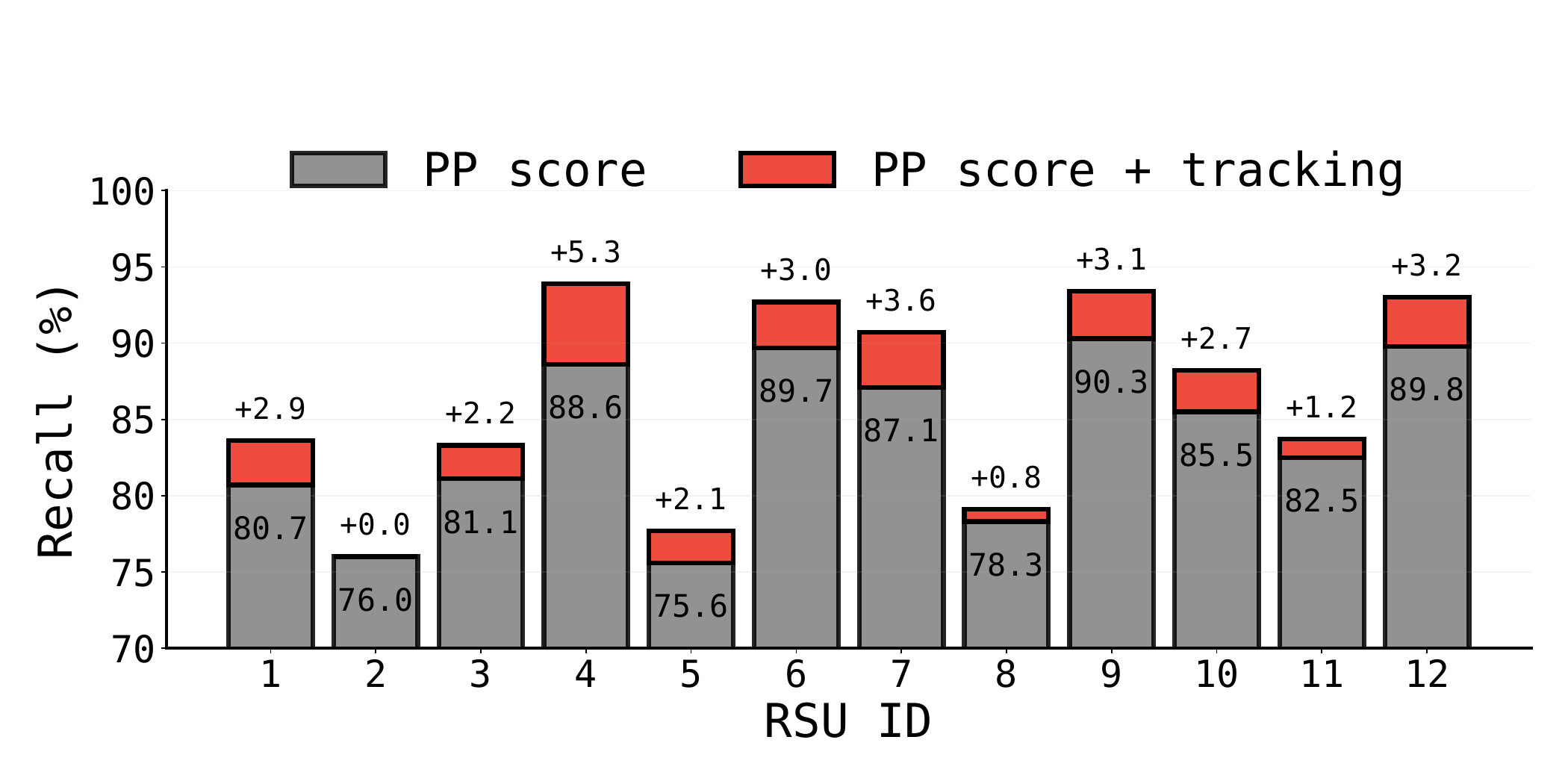}
        \vskip-5pt
        \caption{\textbf{Effect of tracking refinement~\cite{zhang2023oyster}.} Incorporating tracking improves pseudo-label recall, yielding stronger supervision for unsupervised RSU training.}
        \label{fig:tracking}
    \end{minipage}
    % \vspace{-6pt}
\end{figure*}

\subsection{Further Analysis and Insights}

We primarily use CenterPoint~\cite{yin2021centerpoint} as the ego detector in the following analyses.

\mypara{Do static views enable reliable object discovery?}
As shown in~\cref{fig:pp_score}, the fixed viewpoints of RSUs allow them to observe the same scene repeatedly over time, producing highly consistent background points and low-persistence dynamic objects. 
This results in a discriminative PP-score distribution, enabling clean dynamic-object proposals and reliable pseudo-labels for unsupervised training. 
Furthermore, as illustrated in~\cref{fig:tracking}, incorporating tracking consistently improves recall, which further improves pseudo-label quality.

\begin{table*}[t]
\centering

\begin{minipage}[t]{0.57\linewidth}
\centering
\caption{
\textbf{Quality of broadcasted pseudo-labels for ego training in Stage 2.}
Pseudo-labels generally achieve high recall, while communication noise degrades localization quality.
}
\label{tab:stage2_broadcast}
\vskip-5pt
\begin{adjustbox}{width=\linewidth}
\begin{tabular}{l c c c c}
\toprule
 & Comm. & Car & Ped. & Cyc. \\
\cmidrule(lr){3-3}\cmidrule(lr){4-4}\cmidrule(lr){5-5}
Annotation source & Noise & Prec./Rec. & Prec./Rec. & Prec./Rec. \\
\midrule
PP score & $\checkmark$ & 92.4/83.8 & 22.7/19.7 & 29.5/45.3 \\
PP score & -- & 98.3/89.1 & 77.9/67.6 & 54.9/84.3 \\
\cmidrule(lr){1-5}
PP score + tracking & $\checkmark$ & 91.0/83.6 & 23.8/23.6 & 33.9/52.9 \\
PP score + tracking & -- & 97.3/89.5 & 74.3/73.7 & 60.7/94.7 \\
\cmidrule(lr){1-5}
\textcolor{gray}{Supervised RSUs} & \textcolor{gray}{$\checkmark$} & \textcolor{gray}{92.7/94.2} & \textcolor{gray}{23.7/17.2} & \textcolor{gray}{50.8/51.6} \\
\textcolor{gray}{Supervised RSUs} & \textcolor{gray}{--} & \textcolor{gray}{97.2/98.8} & \textcolor{gray}{90.2/65.5} & \textcolor{gray}{93.7/95.2} \\
\bottomrule
\end{tabular}
\end{adjustbox}
\end{minipage}
\hfill
% \begin{minipage}[t]{0.38\linewidth}
% \centering
% \caption{
% \textbf{Adapting a pretrained ego detector using infra-taught pseudo-labels.}
% A detector trained in rural town is evaluated on urban town.
% Fine-tuning with pseudo-labels from new town improves performance under this domain shift.
% }
% \label{tab:ego_pretrain}
% \vskip-5pt
% \begin{adjustbox}{width=\linewidth}
% \begin{tabular}{c c c c c}
% \toprule
% Train town & ~~Car~~ & ~~Ped.~~ & ~~Cyc.~~ & \cellcolor{blue!10}~~Avg.~~\\
% \midrule
% 10 & 82.3 & 79.3 & 68.5 & \cellcolor{blue!10}76.7 \\ \cmidrule(l){1-5}
% 7 & 59.2 & 66.2 & 67.5 & \cellcolor{blue!10}64.3 \\
% 7$\xrightarrow{}$10 & 81.9 & 79.9 & 76.1 & \cellcolor{blue!10}79.3 \\
% \bottomrule
% \end{tabular}
% \end{adjustbox}
% \end{minipage}
\begin{minipage}[t]{0.41\linewidth}
\centering
% \caption{
% \textbf{Adapting a pretrained ego detector using infra-taught pseudo-labels.}
% A detector trained in rural town is evaluated on urban town.
% Fine-tuning with pseudo-labels from new town improves performance under this domain shift.
% }
\caption{
\textbf{Adapting a pretrained ego detector using infra-taught pseudo-labels.}
A detector trained in a rural town is evaluated in an urban town.
Fine-tuning with pseudo-labels from the new town improves performance under this domain shift.
}
\label{tab:ego_pretrain}
\vskip-7pt
\begin{adjustbox}{width=\linewidth}
\begin{tabular}{l c c c c}
\toprule
Train domain & ~~Car~~ & ~~Ped.~~ & ~~Cyc.~~ & \cellcolor{blue!10}~~Avg.~~\\
\midrule
\textcolor{gray}{Urban} & \textcolor{gray}{82.3} & \textcolor{gray}{79.3} & \textcolor{gray}{68.5} & \cellcolor{blue!10}\textcolor{gray}{76.7} \\ \cmidrule(l){1-5}
Rural & 59.2 & 66.2 & 67.5 & \cellcolor{blue!10}64.3 \\
Rural$\xrightarrow{}$Urban & 81.9 & 79.9 & 76.1 & \cellcolor{blue!10}79.3 \\
\bottomrule
\end{tabular}
\end{adjustbox}
\end{minipage}
\vskip-5pt
\end{table*}

\begin{figure}[t]
\centering
\begin{minipage}{0.62\linewidth}
    \centering
    \includegraphics[width=\linewidth]{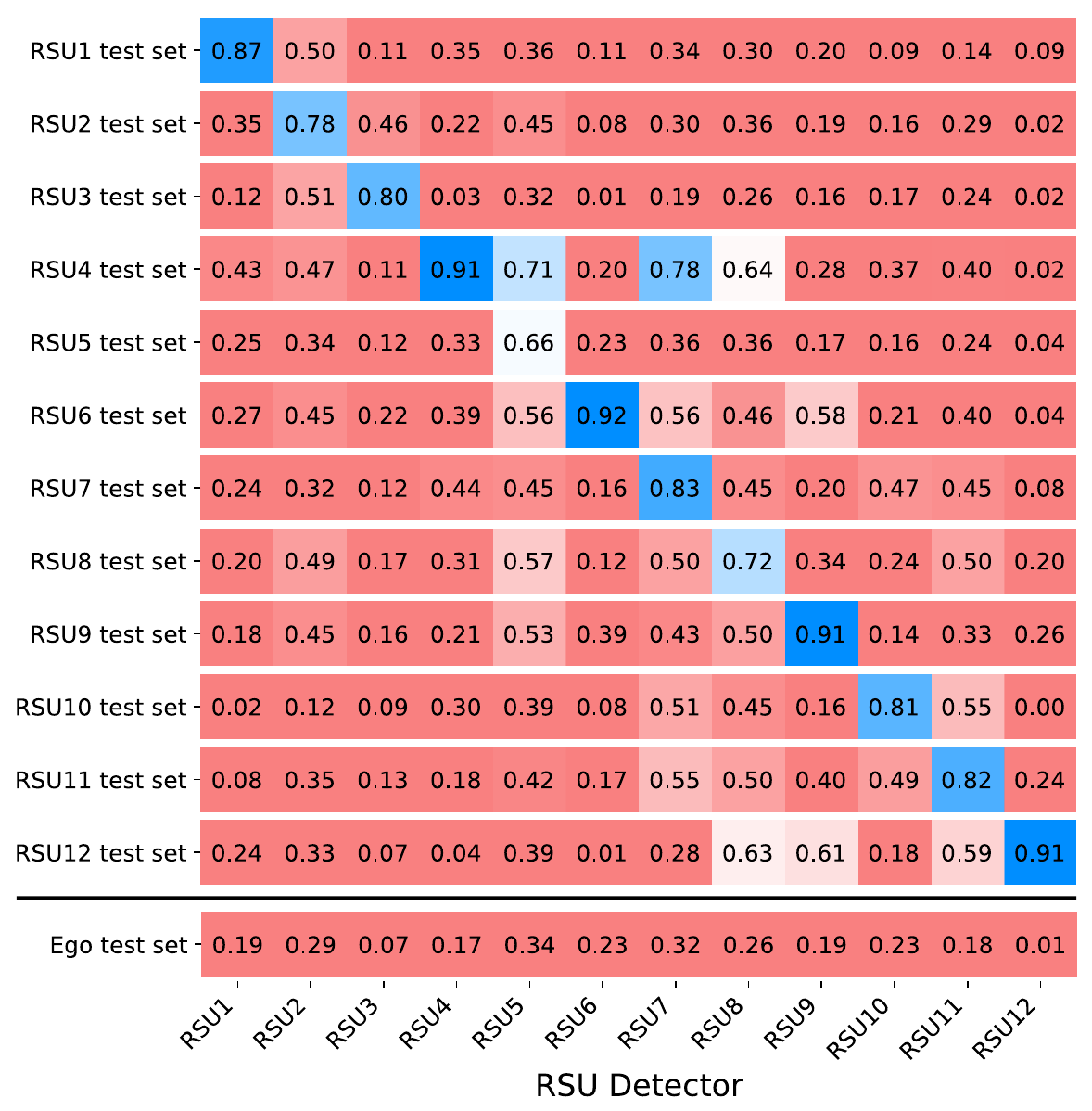}
\end{minipage}
\hfill
\begin{minipage}{0.33\linewidth}
    \caption{
\textbf{Cross-evaluation across RSUs.}
Each RSU detector is trained at its own location and evaluated across all other RSU viewpoints.
The heatmap shows that detectors perform strongly within their own trained view but degrade substantially when evaluated at different locations, indicating that RSU detectors are location-specific.
Furthermore, detectors trained at RSU viewpoints fail to generalize to the ego viewpoint (last row).
}
\label{fig:exp_cross_eval_rsu}
\end{minipage}
% \vskip-20pt
\end{figure}

\mypara{Do RSU detectors generalize beyond their training views?}
To assess whether an RSU trained at a fixed location can generalize beyond its own field of view, we perform cross-evaluation across RSUs. As shown in~\cref{fig:exp_cross_eval_rsu}, each RSU achieves strong performance on its own viewpoint but degrades when evaluated on frames from other locations. 
Moreover, the last row shows that simply applying any individual RSU detector to the \emph{ego viewpoint} does not generalize across the town.
This highlights the inherently local nature of RSU detectors and motivates a distributed teacher composed of multiple complementary RSUs.

\begin{table*}[t]
\centering
\begin{minipage}[t]{0.66\linewidth}
\centering
\caption{
\textbf{Scalability study.}
We synthesize a unified environment by combining data from four towns and train 48 RSU detectors to generate pseudo-labels for ego training. 
A single ego detector is then trained using the aggregated supervision. 
Unlike~\cref{tab:stage3_ego}, evaluation here is performed on the \emph{combined test sets} from all four towns.
}
\label{tab:general_ego}
\vskip-9pt
\begin{adjustbox}{width=\linewidth}
\begin{tabular}{l c c c c c c}
\toprule
Stage 1 (RSU training) & \multicolumn{2}{c}{Stage 2 (Broadcasting)} & \multicolumn{4}{c}{Stage 3 (Ego testing)} \\
\cmidrule(lr){1-1} \cmidrule(lr){2-3} \cmidrule(lr){4-7}
Annotation source & Comm.\ noise & Refine & Car & Ped. & Cyc. & \cellcolor{blue!10}Avg. \\
\midrule

\rowcolor{gray!10} \multicolumn{7}{l}{\emph{Ego detector: PointPillars~\cite{lang2019pointpillars}}} \\

\textbf{48} RSUs (unsup.) & -- & -- & 84.1 & 78.4 & 80.2 & \cellcolor{blue!10}80.9 \\
\textbf{48} RSUs (unsup.) & $\checkmark$ & -- & 80.4 & 56.8 & 77.8 & \cellcolor{blue!10}71.7 \\
\textbf{48} RSUs (unsup.) & $\checkmark$ & $\checkmark$ & 81.7 & 63.9 & 81.7 & \cellcolor{blue!10}75.8 \\

\cmidrule{1-7}

\textcolor{gray}{Ego ground-truth} 
& \textcolor{gray}{--} 
& \textcolor{gray}{--} 
& \textcolor{gray}{92.4} 
& \textcolor{gray}{89.3} 
& \textcolor{gray}{94.6}
& \cellcolor{blue!10}\textcolor{gray}{92.1} \\

\midrule

\rowcolor{gray!10} \multicolumn{7}{l}{\emph{Ego detector: CenterPoint~\cite{yin2021centerpoint}}} \\

\textbf{48} RSUs (unsup.) & -- & -- & 82.7 & 81.0 & 78.3 & \cellcolor{blue!10}80.7 \\
\textbf{48} RSUs (unsup.) & $\checkmark$ & -- & 78.2 & 53.5 & 72.5 & \cellcolor{blue!10}68.1 \\
\textbf{48} RSUs (unsup.) & $\checkmark$ & $\checkmark$ & 80.5 & 66.4 & 81.5 & \cellcolor{blue!10}76.1 \\

\cmidrule{1-7}

\textcolor{gray}{Ego ground-truth} 
& \textcolor{gray}{--} 
& \textcolor{gray}{--} 
& \textcolor{gray}{91.0} 
& \textcolor{gray}{90.6} 
& \textcolor{gray}{95.3}
& \cellcolor{blue!10}\textcolor{gray}{92.3} \\

\bottomrule
\end{tabular}
\end{adjustbox}
\end{minipage}
\hfill
% ---------------- Right table ----------------
\begin{minipage}[t]{0.3\linewidth}
\centering
\caption{
\textbf{Complementarity with ego-centric unsupervised methods.}
Combining our pseudo-labels with existing approaches further improves performance.
Evaluation follows~\cite{you2022modest} with PointRCNN~\cite{shi2019pointrcnn} as the detector. ``Dyn'' denotes dynamic objects.
}
\label{tab:ego_unsup_synergy}
\vskip-8pt
\begin{adjustbox}{width=\linewidth}
\begin{tabular}{l c c}
\toprule
Anno. source & ~~Noise~~ & ~~Dyn.~~ \\
\midrule
MODEST~\cite{you2022modest} & -- & 18.0 \\
Oyster~\cite{zhang2023oyster} & -- & 39.8 \\ \cmidrule(lr){1-3}
Ours & -- & 62.0 \\ 
Ours & $\checkmark$ & 49.9 \\ \cmidrule(lr){1-3}
Oyster + Ours & -- & 64.9 \\
Oyster + Ours & $\checkmark$ & 59.5 \\
\bottomrule
\end{tabular}
\end{adjustbox}
\end{minipage}
\end{table*}

\mypara{Is the paradigm geographically scalable?}
We conduct an experiment to examine whether the proposed pipeline scales to larger environments. 
Since existing datasets do not provide sufficient geographic coverage for such analysis, we combine our data from all four towns and treat them as a \emph{unified} environment. 
In this setup, we train RSU detectors in an unsupervised manner across all 48 RSUs (12 per town) using PP-score proposals with tracking, and use their predictions to broadcast pseudo-labels for ego training. 
A single ego detector is then trained using the aggregated supervision and evaluated on the test splits of all towns. 
As shown in~\cref{tab:general_ego}, the results exhibit trends similar to those observed in~\cref{tab:stage3_ego}, suggesting that the proposed paradigm can naturally extend to larger-scale deployments.

\begin{figure*}[t]
    \centering
    \begin{minipage}[t]{0.48\linewidth}
        \centering
        \includegraphics[width=\linewidth]{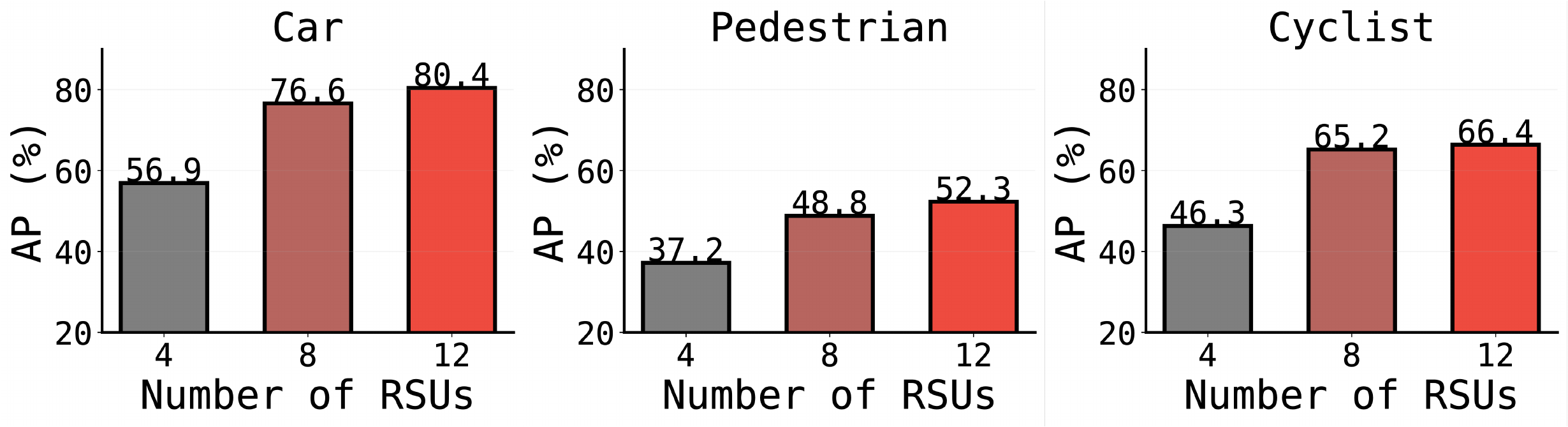}
        \vskip-5pt
        \caption{\textbf{Effect of the number of RSUs.} Increasing the number of available RSUs improves ego detector performance by providing more diverse and complementary pseudo-label supervision.}
        \label{fig:exp_num_rsu}
    \end{minipage}\hfill
    \begin{minipage}[t]{0.48\linewidth}
        \centering
        \includegraphics[width=\linewidth]{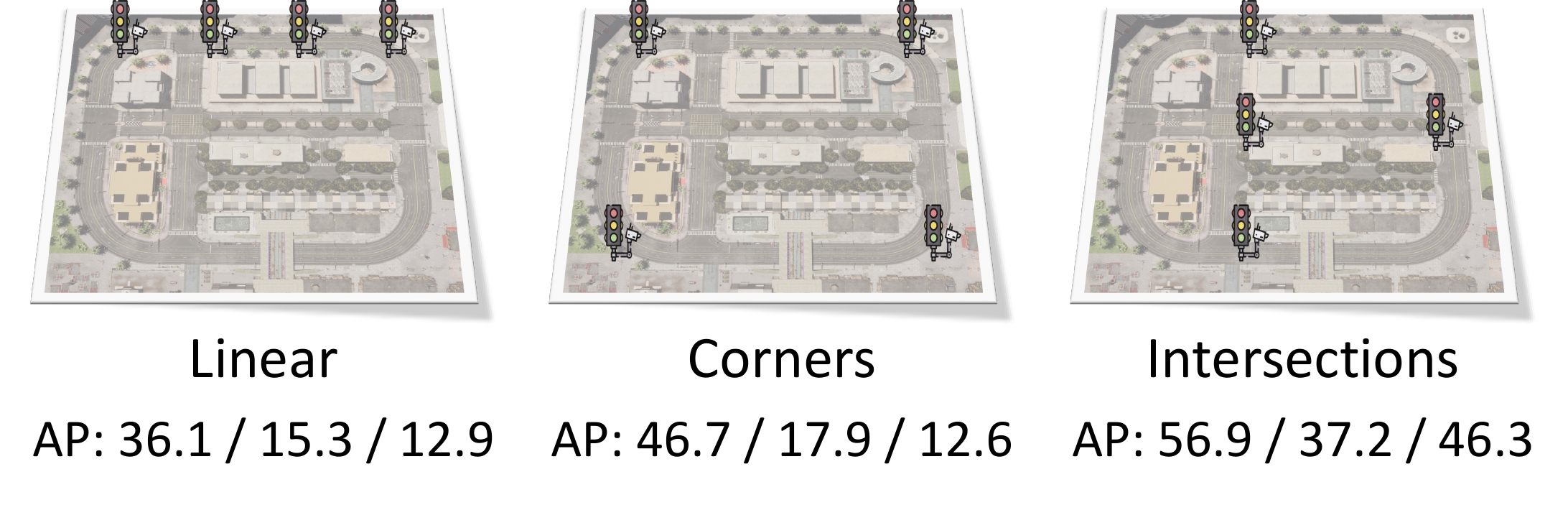}
        \vskip-5pt
        \caption{\textbf{Effect of RSU placement.} Intersection placement provides more balanced coverage and higher-quality pseudo-labels than linear or corner layouts. APs: \textit{Car}, \textit{Pedestrian}, and \textit{Cyclist}.}
        \label{fig:exp_rsu_placement}
    \end{minipage}
    % \vskip-15pt
\end{figure*}

\mypara{Does increasing the number of RSUs help?}
We investigate how the number of RSUs affects ego detector performance within our experimental setting. 
For fair comparison, we collect additional data for configurations with 4 and 8 RSUs while keeping the total number of training frames the same as the 12-RSU setup. 
As shown in~\cref{fig:exp_num_rsu}, performance improves as the ego receives pseudo-labels from more RSUs. 
This trend suggests that additional RSUs provide more diverse and complementary supervision during ego training.

\mypara{Where should RSUs be placed?}
We next examine how RSU placement affects ego detector performance under a limited-budget scenario where only four RSUs can be deployed. 
Unlike~\cite{jiang2023optimizing}, which studies optimal placement for \emph{online} multi-RSU collaborative perception, our focus is on how placement influences the \emph{offline} training quality of the ego detector. 
For fair comparison, we collect training data for each placement strategy while keeping the total number of training frames the same. 
As shown in~\cref{fig:exp_rsu_placement}, we compare three layouts: \emph{linear}, \emph{corner}, and \emph{intersection}. 
Among these, intersection placement yields the strongest performance, suggesting that locations with high traffic flow provide richer pseudo-labels and more diverse supervision for ego training.

\mypara{Does the pipeline help adapt detectors across domains?}
We study whether infrastructure-taught pseudo-labels can adapt an ego detector trained in a different geographic domain. 
Specifically, we take CenterPoint~\cite{yin2021centerpoint} trained on a rural town (Town~7) and evaluate it directly on an urban town (Town~10). As shown in~\cref{tab:ego_pretrain}, the model performs poorly under this domain shift (\eg, only 59.2\% Car AP).
We note that this setting is intentionally designed to induce a meaningful domain shift.
Fine-tuning the model previously trained in a rural environment with infrastructure-provided pseudo-labels from the current urban town leads to consistent improvements, and in some cases even surpasses training the ego detector from scratch (\eg, Cyclist AP increases to 76.1\%, compared to 68.5\%). These results indicate that infra-taught supervision can potentially serve as an effective signal for \emph{label-free domain adaptation}. 

\mypara{Can our pseudo-labels complement ego-centric pseudo-labels?}
Although our goal is not to compete with existing ego-centric unsupervised methods, but rather to explore a new source of supervision, we compare the effectiveness of our pseudo-labels with representative ego-centric approaches: MODEST~\cite{you2022modest} and Oyster~\cite{zhang2023oyster}. 
To isolate the effect of the supervision source, we disable self-training for all methods and tune hyperparameters to the best of our ability. 
As shown in~\cref{tab:ego_unsup_synergy}, training an ego detector using our pseudo-labels alone already outperforms the baselines. 
More importantly, combining both sources (details in \textcolor{red}{Suppl. Sec. C.3}) yields further improvements, indicating that infrastructure-taught pseudo-labels and ego-centric learning provide \emph{complementary signals}.

\section{Conclusion and Discussion}

We formulate \emph{infrastructure-taught, label-free} 3D perception, a new paradigm in which stationary RSUs learn from their own observations and supervise ego vehicles. 
Through a controlled concept-and-feasibility study, we show that distributed RSUs can provide useful supervision for training standalone ego detectors, highlighting the potential for city infrastructure itself to provide a \emph{scalable supervisory signal} for autonomous vehicles.
We hope this work provides a concrete foundation for further exploration of infrastructure-taught ego learning.

\mypara{Future Opportunities.}
Our study opens several promising directions for future work. 
First, the current label-free RSU training primarily focuses on dynamic objects by leveraging temporal cues from traffic scenes. 
Extending the paradigm to static categories is an important next step and may require incorporating additional signals such as scene-level context.
Second, while our current pipeline focuses on LiDAR-based detection, integrating multimodal inputs (\eg, camera and LiDAR) could further improve the robustness of infrastructure-taught supervision.
Third, our experiments are conducted in a controlled simulation environment, which enables systematic analysis of the proposed paradigm. A natural next step is to validate the pipeline in real-world deployments as connected infrastructure becomes increasingly available.

% \newpage

\section*{Acknowledgements}
This research is supported in part by grants from the National
Science Foundation (IIS-2107077, IIS-2107161, CNS-2211599, IIS-2305532).
We thank the Ohio Supercomputer Center for the generous support
of the computational resources.

\bibliographystyle{splncs04}
\bibliography{main}

\clearpage
\setcounter{page}{1}
% \maketitlesupplementary

\renewcommand{\thesection}{\Alph{section}}
\renewcommand{\thesubsection}{\thesection.\arabic{subsection}}
\renewcommand{\thetable}{A\arabic{table}}
\renewcommand{\thefigure}{A\arabic{figure}}
\renewcommand{\theequation}{A\arabic{equation}}
\setcounter{figure}{0}
\setcounter{table}{0}
\setcounter{equation}{0}
\setcounter{section}{0}

In this supplementary material, we provide additional details and experimental results that complement the main paper. Specifically, we include:

\begin{itemize}
\item \cref{sec:additional_discussion}: Disclosure of LLM usage.
\item \cref{sec:additional_details_on_civet}: Further details on the curated dataset.
\item \cref{sec:additional_implementation_details}: Additional implementation details.
\item \cref{sec:additional_experimental_results}: Expanded experimental results.
\end{itemize}

\section{Disclosure of LLM Usage}
\label{sec:additional_discussion}

Portions of this manuscript were edited for clarity and readability using a large language model (LLM).
The LLM was not used to generate research ideas, design experiments, analyze data, or draw conclusions.
All scientific content, methodologies, and results are solely the authors’ original work.

\section{Additional Details on \textsc{CIVET}}
\label{sec:additional_details_on_civet}

\subsection{Data Curation}

\begin{figure}
    \centering
    \includegraphics[width=\linewidth]{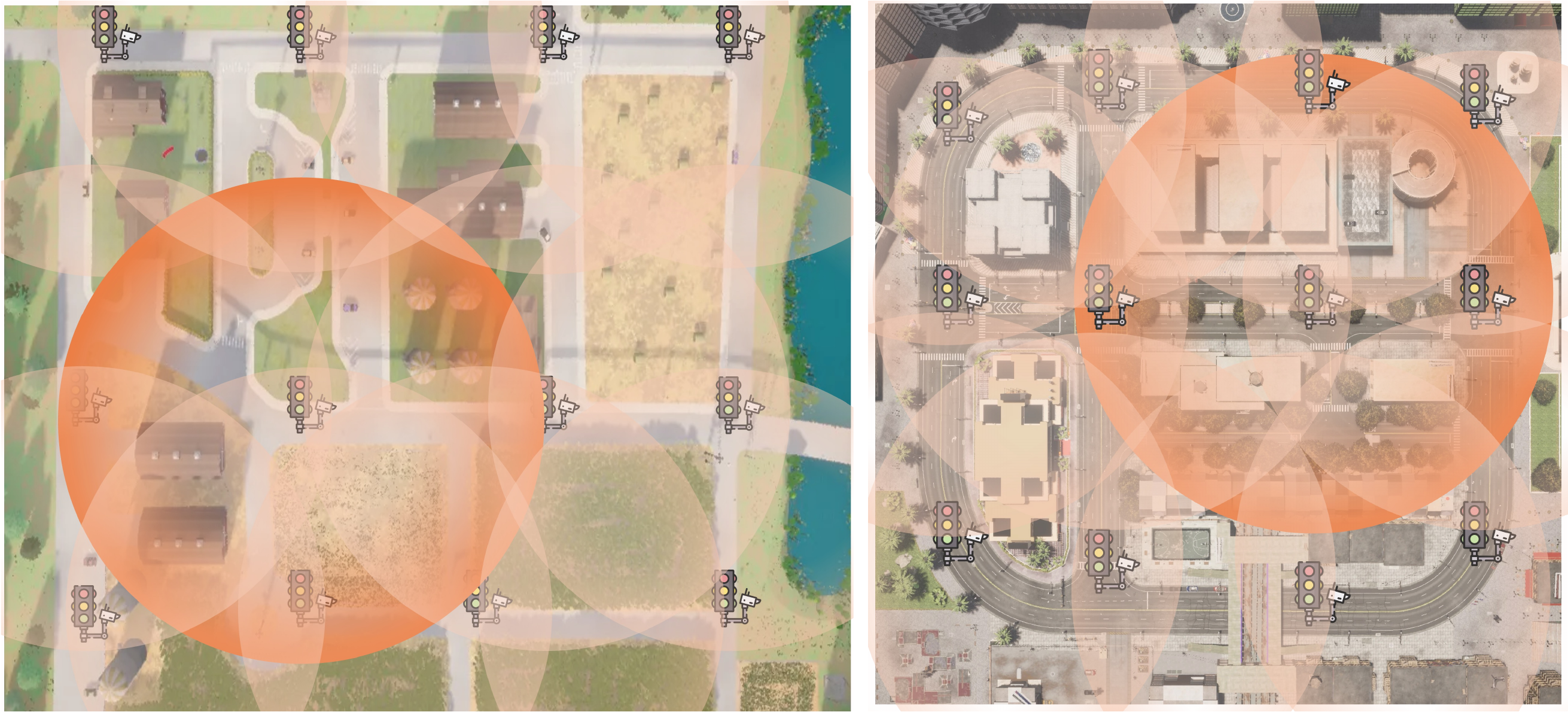}
    \caption{
\textbf{RSU deployment and coverage in Town~7 and Town~10.}
Each shaded region denotes the field of view of an RSU.
Together, the RSUs provide complementary coverage of the geo-fenced
environment, enabling infrastructure-taught supervision across
most of the drivable area. One region is shaded slightly darker for visualization.
}
\vskip-10pt
\label{fig:rsu_deployment}
\end{figure}

\begin{figure}
    \centering
    \vskip-5pt
    \includegraphics[width=\linewidth]{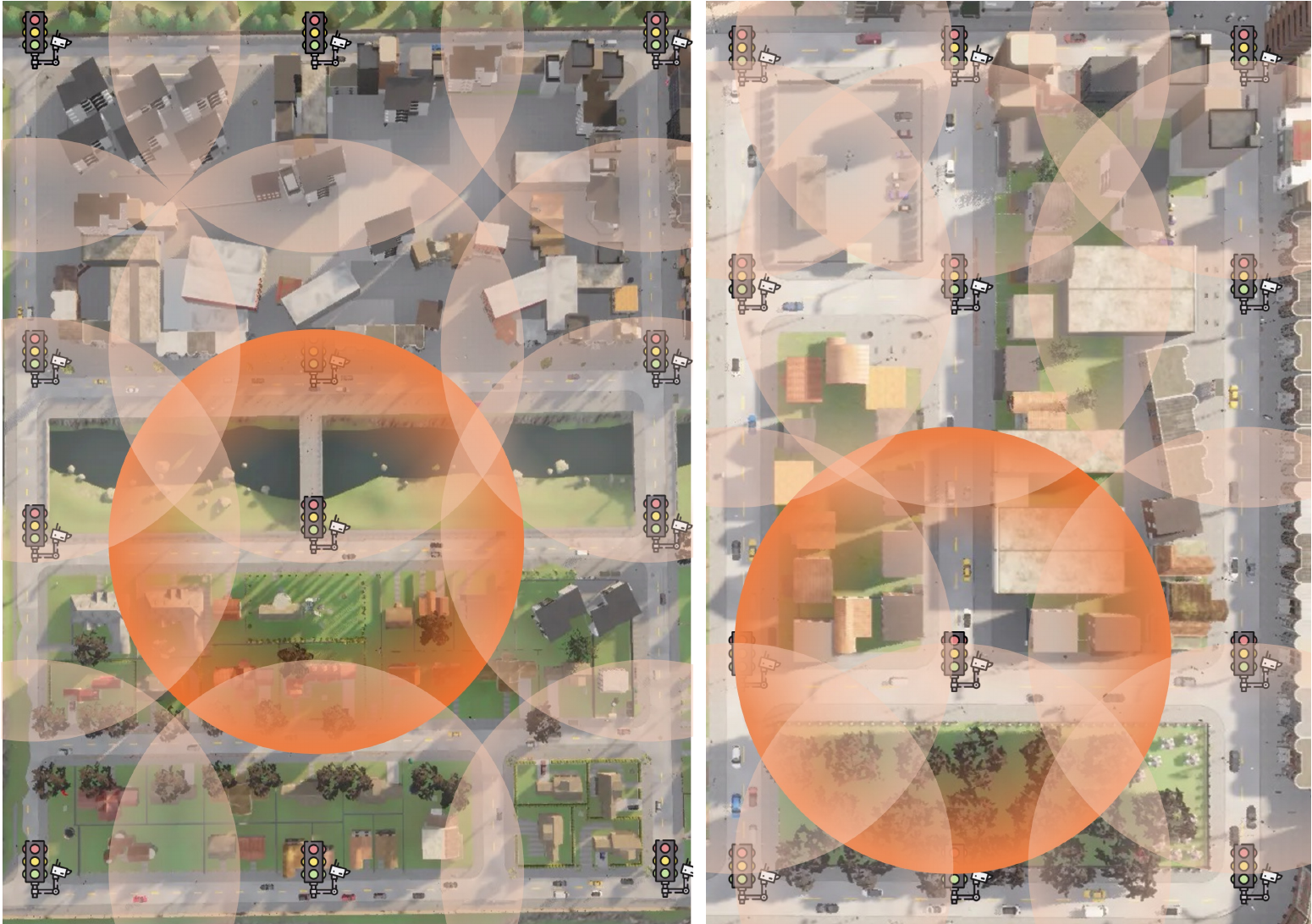}
    \caption{\textbf{RSU deployment and coverage in Town~1 and Town~2.} One region is shaded slightly darker for clearer visualization.}
    \label{fig:rsu_deployment_01_02}
\end{figure}

\mypara{Town environments.}
Our main study focuses on Town~7 and Town~10, which capture distinct rural and urban driving conditions, differing in road layout, traffic density, and vehicle composition (\cref{tab:town_stats}). 
Town~7 represents a low-density rural setting with larger vehicles, whereas Town~10 features dense urban traffic with smaller and more diverse actors. 
The RSU deployment for both towns is shown in~\cref{fig:rsu_deployment}, where each RSU's field of view is visualized; one coverage region is shaded slightly darker for clarity. 
As illustrated, the RSUs collectively cover most of the drivable area, enabling comprehensive supervision throughout the geo-fenced space.
In addition, we collect data from Town~1 and Town~2 to study scalability. 
The RSU placement for these environments is shown in~\cref{fig:rsu_deployment_01_02}.

\subsubsection{World initialization.}
We establish a CARLA client and initialize the simulator in synchronous mode with a fixed simulation time step of 1/20 seconds (20\,Hz). 
The Traffic Manager is also configured in synchronous mode to ensure deterministic behavior of background traffic. 

\subsubsection{Route and scenario loading.}
For each simulation route, the system first loads the target map and then configures traffic-light behavior depending on the stage of the pipeline. 
In Stage~1, all traffic lights are disabled to prevent prolonged stopping at intersections, ensuring continuous traffic flow.
In Stage~2 and Stage~3, normal traffic-light control is enabled, introducing realistic driving patterns and dynamics for RSU-to-ego broadcasting and ego evaluation.

\subsubsection{Agent initialization.}
\begin{figure}
    \centering
    \vskip-5pt
    \includegraphics[width=\linewidth]{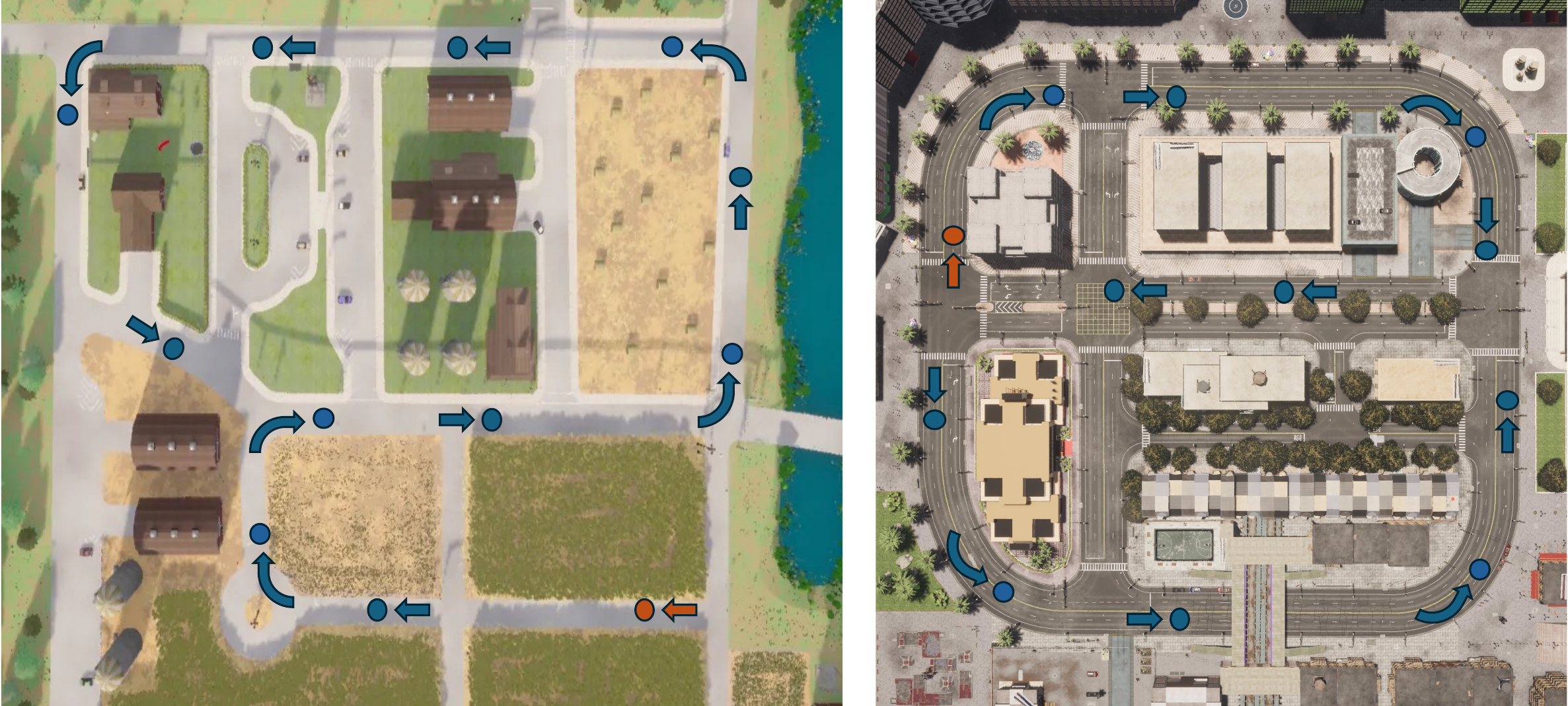}
    \caption{\textbf{Reference driving paths in Town~7 and Town~10.} 
The ego vehicle follows high-level reference routes designed to traverse most of the drivable area, ensuring broad coverage during data collection.}
    \label{fig:suppl_path_town_7_10}
\end{figure}

\begin{figure}
    \centering
    \vskip-5pt
    \includegraphics[width=\linewidth]{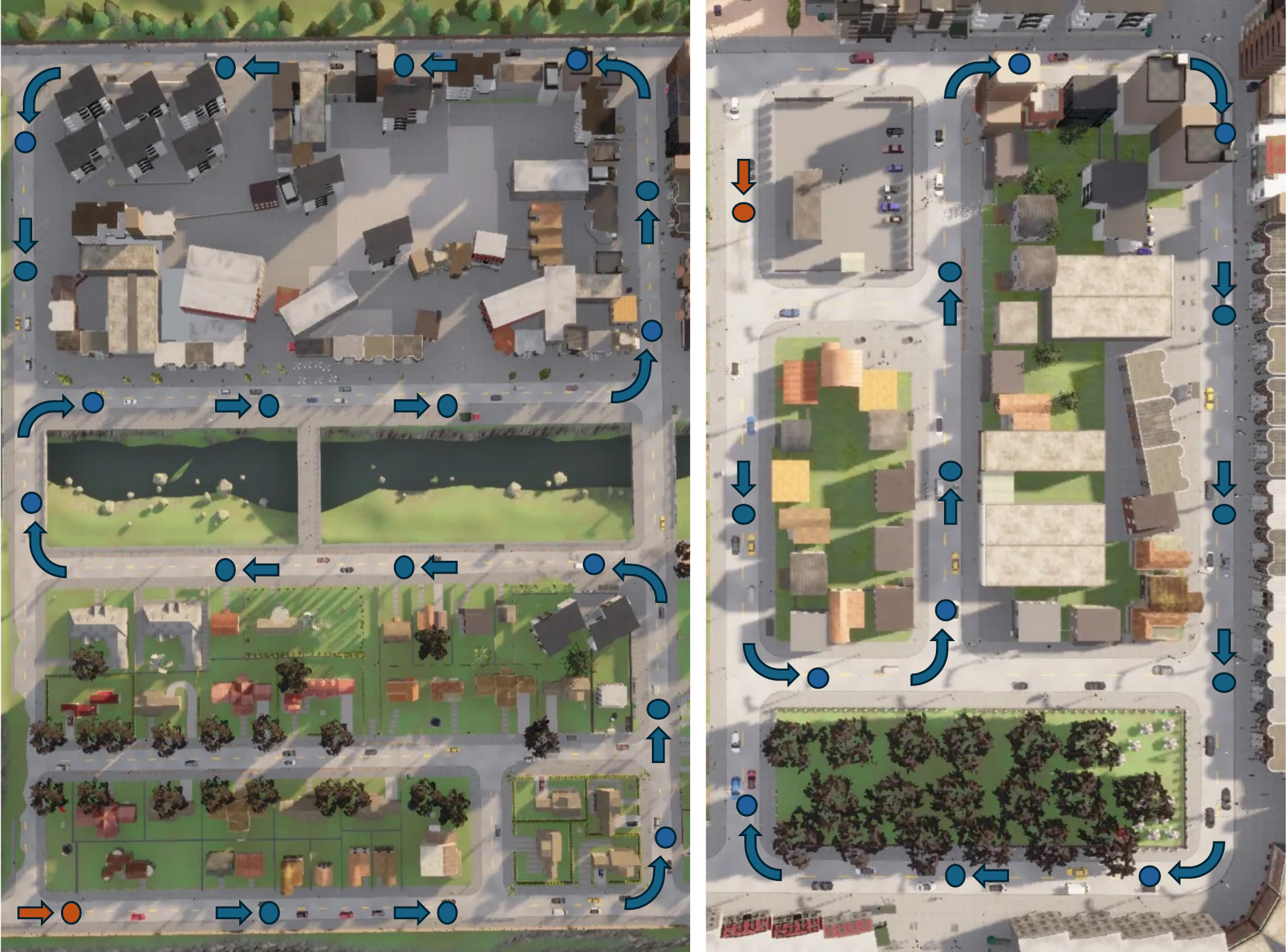}
    \caption{\textbf{Reference driving paths in Town~1 and Town~2.} }
    \label{fig:suppl_path_town_1_2}
\end{figure}

The ego vehicle is initialized at a fixed starting pose and is assigned a high-level reference route for each scenario (\cref{fig:suppl_path_town_7_10} and~\cref{fig:suppl_path_town_1_2}). 
Note that this route specifies the intended global path but does not prescribe fine-grained control. As a result, the ego’s actual trajectory varies across simulation runs due to interactions with surrounding agents and dynamic traffic conditions.
In contrast, all non-ego actors---including cars, pedestrians, and cyclists---are randomly spawned throughout the map at the beginning of each episode. Their motion plans are sampled from CARLA’s navigation graph, introducing diverse and unpredictable traffic behaviors. This randomized initialization ensures that, even with the same high-level route, the ego experiences different encounters and visibility patterns in every run, increasing the diversity of collected data and mitigating overfitting to any single traffic configuration.

\subsubsection{Closed-loop scenario execution.}
At each simulation tick, the system runs in a fully closed-loop setup. All ego onboard sensors (RGB cameras, LiDAR, GNSS, IMU) and all 12 RSUs (mounted at 7.5\,m) produce synchronized observations, which are delivered through CARLA’s synchronous \texttt{SensorInterface}. The autonomous agent fuses ego-vehicle and RSU observations and outputs control commands (steering, throttle, brake).
Although observations are generated every tick, we record them at a downsampled rate of once every two frames to reduce storage and computation.

\subsubsection{Environment and vehicle assets.}
We intentionally differentiate the domains of Town~7 and Town~10 to study the general applicability of our proposed paradigm.
Detailed environmental statistics of Town~7 and Town~10 are summarized in~\cref{tab:town_stats}. Town~7 features larger vehicles and sparser traffic, whereas Town~10 features smaller urban vehicles and a higher density of pedestrians and cyclists. The physical dimensions of all vehicle models are provided in~\cref{tab:model_sizes_presence}.
Each town uses a distinct subset of vehicles to reflect realistic rural and urban traffic patterns—for example, Town~7 includes sedans, pickups, and trucks, while Town~10 primarily uses compact cars and microcars suited for dense urban streets.

\mypara{Overall metadata.}
Each ego frame includes synchronized RGB, depth, LiDAR, semantic LiDAR, segmentation maps, bird’s-eye views, camera parameters, and navigation states such as speed, waypoints, and IMU. RSU frames contain multi-view RGB, depth, high-resolution LiDAR, semantic LiDAR, and calibration data. These modalities are retained for all collected frames and used to support different stages of the pipeline and subsequent analysis.

\begin{table}[ht]
\centering
\caption{
\textbf{Comparison of environment statistics between the two towns.}
The rural town contains larger vehicles and fewer pedestrians/cyclists, whereas the urban town features smaller vehicles and denser non-vehicle traffic. 
Actor counts are reported as car/pedestrian/cyclist, and the average car size is given in (length $\times$ width $\times$ height).
}
\label{tab:town_stats}
\begin{adjustbox}{width=.5\linewidth, center}
\begin{tabular}{ccc}
\toprule
Scene type & \# Actors & ~~~~~~~Avg. car size~~~~~~~ \\
\midrule
Rural & 50 / 25 / 25 & 5.22 $\times$ 2.18 $\times$ 1.72 \\
Urban & 50 / 100 / 50 & 3.99 $\times$ 1.89 $\times$ 1.59 \\
\bottomrule
\end{tabular}
\end{adjustbox}
\end{table}

\begin{table}[ht]
\centering
\caption{\textbf{Vehicles and their dimensions for each town.} $\checkmark$ indicates that the vehicle model is used in the corresponding town.}
\label{tab:model_sizes_presence}
\begin{adjustbox}{width=.6\linewidth}
\begin{tabular}{lccccc}
\toprule
Vehicle & L (m) & W (m) & H (m) & Rural & Urban \\
\midrule
mercedesccc (\textbf{ego}) & 4.67 & 2.00 & 1.44 & $\checkmark$ & $\checkmark$ \\ \cmidrule(lr){1-6}
chevrolet.impala  & 5.36 & 2.03 & 1.41 & $\checkmark$ & -- \\
chargercop2020    & 5.24 & 2.10 & 1.64 & $\checkmark$ & -- \\
benz.coupe        & 5.03 & 2.15 & 1.64 & $\checkmark$ & -- \\
charger2020       & 5.01 & 2.10 & 1.53 & $\checkmark$ & -- \\
dodge.charger.police & 4.97 & 2.04 & 1.55 & $\checkmark$ & -- \\
tesla.cybertruck  & 6.27 & 2.39 & 2.10 & $\checkmark$ & -- \\
carlacola         & 5.20 & 2.61 & 2.47 & $\checkmark$ & -- \\
\cmidrule(lr){1-6}
audi.tt           & 4.18 & 1.99 & 1.39 & -- & $\checkmark$ \\
seat.leon         & 4.19 & 1.82 & 1.47 & -- & $\checkmark$ \\
toyota.prius      & 4.51 & 2.01 & 1.52 & -- & $\checkmark$ \\
nissan.patrol     & 4.60 & 1.93 & 1.85 & -- & $\checkmark$ \\
volkswagen.t2     & 4.48 & 2.07 & 2.04 & -- & $\checkmark$ \\
citroen.c3        & 3.99 & 1.85 & 1.62 & -- & $\checkmark$ \\
mini.cooperst     & 3.81 & 1.97 & 1.48 & -- & $\checkmark$ \\
audi.a2           & 3.71 & 1.79 & 1.55 & -- & $\checkmark$ \\
nissan.micra      & 3.63 & 1.85 & 1.50 & -- & $\checkmark$ \\
bmw.isetta        & 2.21 & 1.48 & 1.38 & -- & $\checkmark$ \\
jeep.wrangler.rubicon & 3.87 & 1.91 & 1.88 & -- & $\checkmark$ \\
\bottomrule
\end{tabular}
\end{adjustbox}
\end{table}

\subsection{Sample Data}

\begin{figure*}
    \centering
    \includegraphics[width=\linewidth, height=\textheight, keepaspectratio]{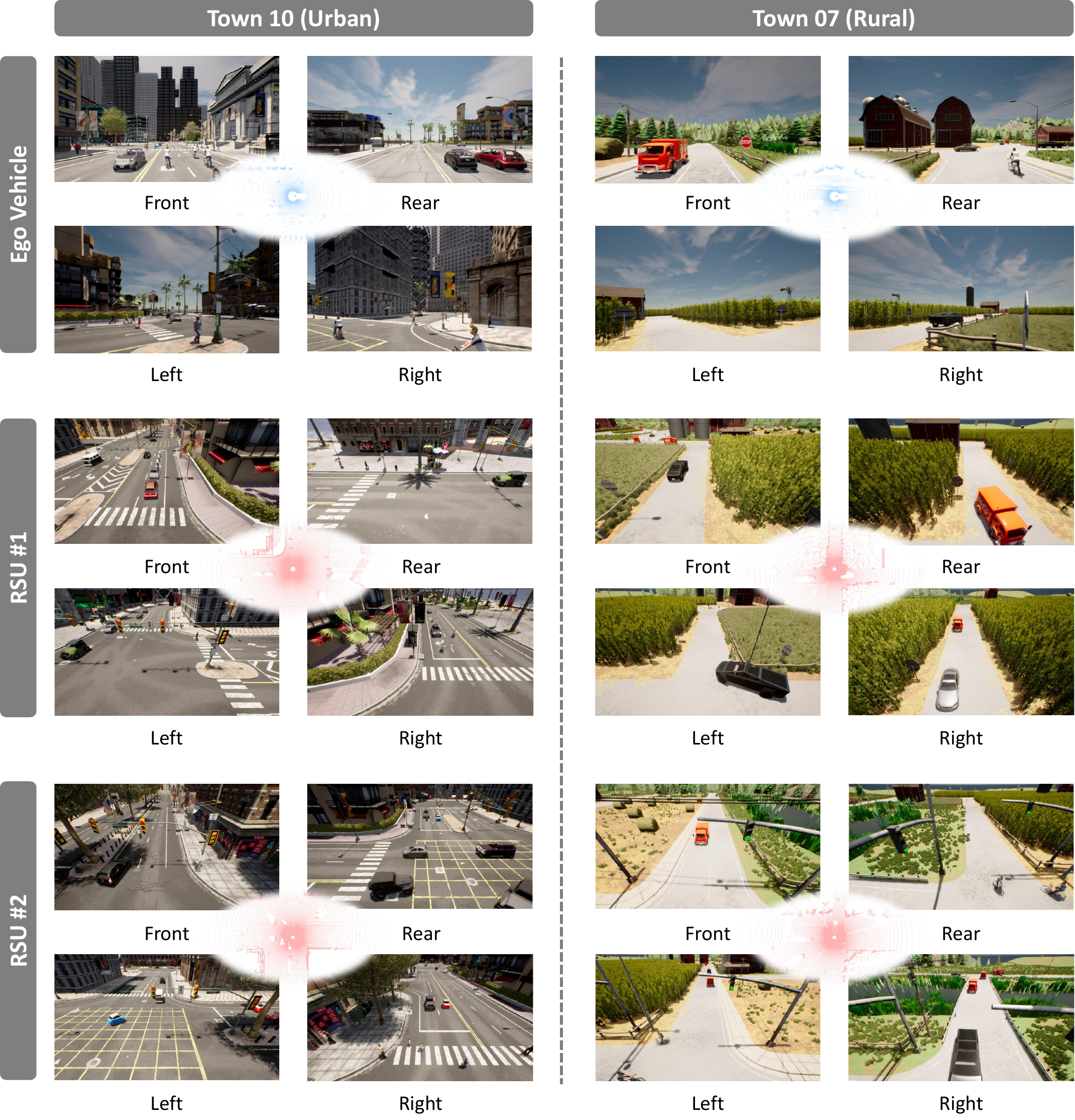}
    \vskip-5pt
    \caption{\textbf{Sample data from the \textsc{CIVET} dataset (urban and rural towns).} 
Each column shows paired examples from the urban and rural environments. 
The two towns exhibit distinct semantics---urban scenes contain smaller vehicles and denser pedestrian/cyclist activity, 
while rural scenes feature larger vehicles and sparser non-vehicle traffic.}
    \label{fig:suppl_sample_data_town_7_10}
\end{figure*}

\begin{figure*}
    \centering
    \includegraphics[width=\linewidth, height=\textheight, keepaspectratio]{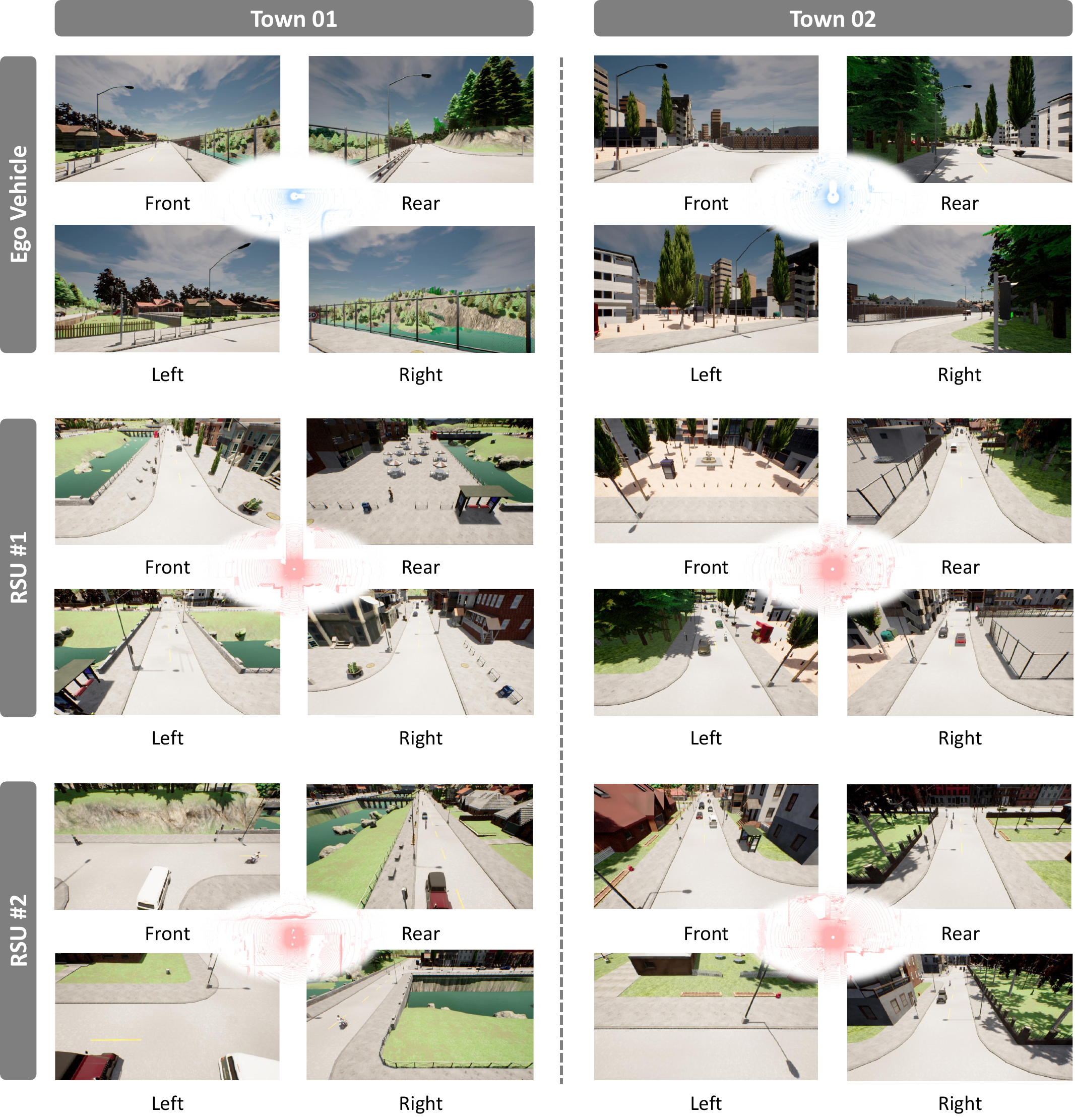}
    \vskip-5pt
    \caption{\textbf{Sample data from the \textsc{CIVET} dataset (additional towns).}
These environments are used to evaluate the scalability of the proposed framework.
}
    \label{fig:suppl_sample_data_town_1_2}
\end{figure*}

\cref{fig:suppl_sample_data_town_7_10} presents samples from our curated dataset, with each column showing paired examples from the urban and rural environments.
As illustrated, the two towns exhibit clearly distinct visual and semantic characteristics: urban scenes contain smaller vehicles and denser pedestrian/cyclist traffic, whereas rural scenes feature larger vehicles and sparser non-vehicle participants. \cref{fig:suppl_sample_data_town_1_2} shows sample scenes from Town~1 and Town~2 used for the scalability study.

\section{Additional Implementation Details}
\label{sec:additional_implementation_details}

\subsection{Unsupervised RSU Training}
Due to computational constraints, we did not perform dedicated hyperparameter tuning for each unsupervised RSU detector. For pseudo-label generation using MODEST~\cite{you2022modest}, we first calculate the PP score using a neighborhood radius of 0.01 across the frames. We then adopt DBSCAN clustering with a neighborhood radius of 0.3 and require at least 2 points to form a valid cluster.
After clustering, we apply common-sense filtering rules: each box must contain at least 5 points, its volume must fall between 0.5 and 20, and its vertical extent must be reasonable---specifically, the maximum height must be at least 0.5\,m and the minimum height must not exceed 2.5\,m.
For tracking refinement, we use the official module provided in~\cite{wu2024cpd}.
While our study uses a unified set of hyperparameters across all RSUs, we expect that tuning these values individually for each RSU could further improve the quality of unsupervised detection. We leave such fine-grained optimization to future work.

\subsection{Broadcasted Pseudo-Label Refinement}
In real-world deployments, communication imperfections during RSU prediction broadcasting---such as localization drift or temporal asynchrony---can introduce noise into the transmitted pseudo-labels.
To mitigate these effects, we adopt a heuristic-based box refinement method~\cite{luo2023drift} as a simple baseline module.
Specifically, for each broadcast pseudo-label box, we generate a small set of candidate boxes by translating the box center over a discrete grid of horizontal offsets \((\Delta x,\Delta y)\) while keeping its size and orientation unchanged.
Each candidate is scored based on the number of LiDAR points that fall inside the box as well as the geometric tightness of the enclosed points.
The candidate with the highest score is selected as the refined box for each actor.
If the support is too weak---\ie, the candidate contains fewer than a predefined minimum number of LiDAR points---we retain the original broadcast box instead of replacing it.

\subsection{Combining Ego-Centric and Infrastructure Pseudo-Labels}

Broadcasted pseudo-labels from infrastructure may occasionally exhibit inaccurate sizes or orientations due to sparse LiDAR observations at long range or communication noise during broadcasting.
In contrast, ego-centric pseudo-labeling methods~\cite{you2022modest,zhang2023oyster} can observe denser LiDAR measurements when objects are closer to the ego vehicle and are not affected by communication noise.
As a result, ego-centric approaches may produce more accurate bounding boxes in such cases.
Motivated by this insight, we combine pseudo-labels from both sources when generating the pseudo-label supervision used for Tab.~5 in the main paper.
For each infrastructure-provided pseudo-label, we check whether an overlapping box exists in the ego-centric pseudo-label set.
If both candidates are present, we select the one containing a larger number of ego LiDAR points, which typically indicates better alignment with the ego observation.
If no overlapping ego-centric pseudo-label exists, the infrastructure-provided pseudo-label is retained.
We see that this simple fusion strategy can effectively leverage the complementary strengths of infrastructure-based supervision and ego-centric observations.

\subsection{RSU Placement for the Number of RSU Analysis}

\begin{figure}
    \centering
    \includegraphics[width=.42\linewidth]{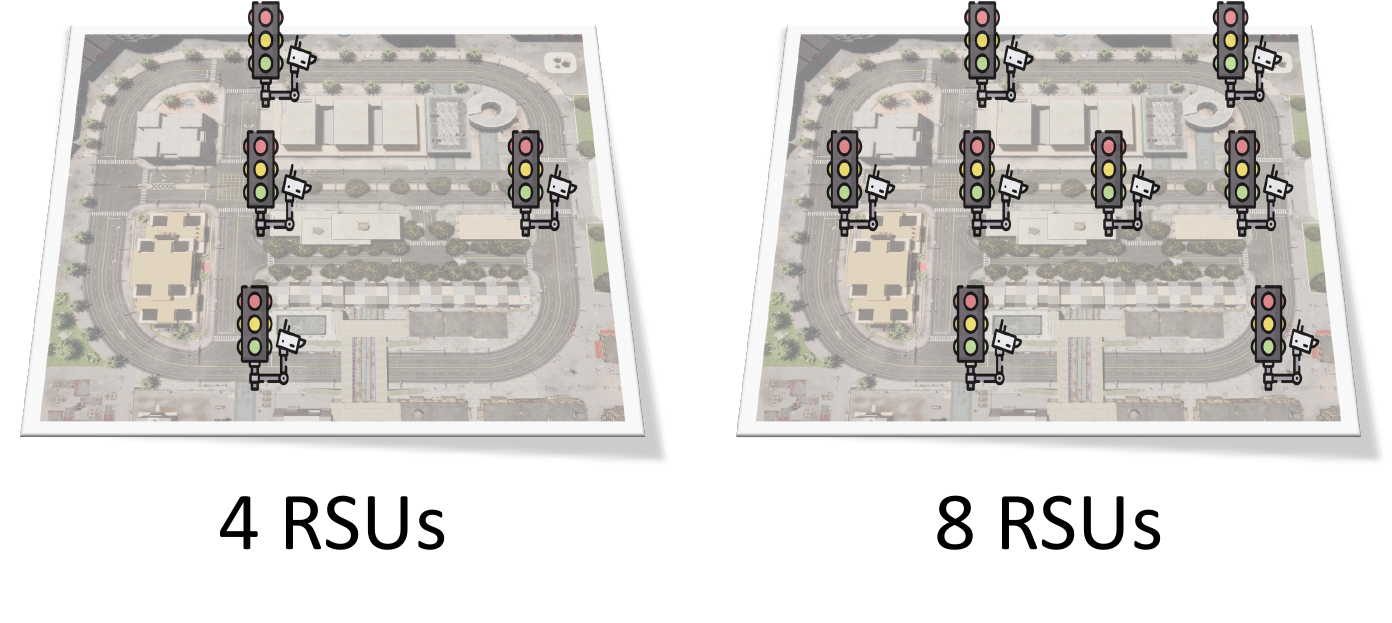}
    \vskip-10pt
    \caption{\textbf{RSU placement used in the analysis of the number of RSUs.}}
    \label{fig:suppl_rsu_placement}
\end{figure}

\cref{fig:suppl_rsu_placement} provides 
RSU placement for Fig.~7 in our study in the main paper.

\section{Additional Experimental Results}
\label{sec:additional_experimental_results}

\subsection{Unsupervised RSU Training}

\mypara{Pseudo-label quality across depth ranges.}
To better understand the quality of infrastructure-provided pseudo-labels, we analyze RSU detection performance as a function of the distance to the RSU.
Specifically, we break down RSU AP (for all classes) by distance-to-RSU, averaged over all 12 RSUs.
As shown in~\cref{tab:rebuttal_pl_quality}, pseudo-label quality is higher for objects located closer to the RSU sensor.
This trend is expected, as nearby objects are observed with denser LiDAR measurements and suffer less from sparsity.
These observations suggest that incorporating distance-aware strategies when aggregating infrastructure-generated pseudo-labels could further improve the supervision quality, which we consider a promising direction for future work.

\begin{table}[t]
\centering
\small
\caption{\textbf{Pseudo-label quality across RSU distance ranges.}
We report RSU detection AP averaged over all RSUs.
Pseudo-label quality is higher for objects closer to the RSU sensor.}
\label{tab:rebuttal_pl_quality}
\begin{adjustbox}{width=.35\linewidth}
\begin{tabular}{lcc}
\toprule
~~Range~~ & ~~0--30m~~ & ~~30--80m~~ \\
\midrule
~~AP & 86.5 & 67.9 \\
\bottomrule
\end{tabular}
\end{adjustbox}
\end{table}

\mypara{Can RSU detector be shared?}
In the main paper, we assume that each RSU is trained independently using its own local observations, without sharing data across different sensors.
This assumption reflects a realistic deployment scenario in which infrastructure sensors operate independently.
However, in practice, it may be possible for RSUs to share their data, enabling the training of a \emph{shared} detector across multiple RSUs.
To investigate this possibility, we train a single detector using the combined data from all RSUs and compare it with detectors trained independently for each RSU.
As shown in~\cref{tab:suppl_shared_rsu}, the shared detector performs slightly better than independently trained RSU detectors.
This result suggests that a general RSU detector is feasible \emph{when data collected from multiple RSUs can be combined together}.

\begin{table}[t]
\centering
\small
\caption{
\textbf{Shared vs.\ independent RSU detectors.}
We compare detectors trained independently for each RSU with a single detector trained using data from all RSUs.
The shared detector performs slightly better, suggesting that a general RSU model is feasible \emph{when data from multiple RSUs can be combined}.
}
\label{tab:suppl_shared_rsu}
\begin{adjustbox}{width=.38\linewidth}
\begin{tabular}{lcc}
\toprule
~~RSU~~ & ~~Independent~~ & ~~Shared~~ \\
\midrule
~~AP & 84.5 & 85.5 \\
\bottomrule
\end{tabular}
\end{adjustbox}
\end{table}

\subsection{Ego Detector Training}

\mypara{Different distance ranges.}
To further understand the effect of infrastructure-taught supervision, we analyze ego detector performance across different distance ranges. 
We use CenterPoint~\cite{yin2021centerpoint} trained on the urban town under the ideal communication setting and evaluate Car AP at IoU~0.5 for near (0--30\,m), mid-range (30--50\,m), and far (50--80\,m) distances. 
As shown in~\cref{tab:depth_range}, our method maintains strong performance even for distant objects, recovering a substantial fraction of the supervised upper bound. 
This highlights an important advantage of infrastructure-taught supervision: RSUs can provide pseudo-labels for objects that are far away or poorly visible from the ego viewpoint, enabling more complete training coverage.

\begin{table}[ht]
\centering
\caption{
\textbf{Ego detector performance across distance ranges.}
Reported numbers represent Car AP@0.5 for CenterPoint~\cite{yin2021centerpoint} on the urban town under noise-free communication.  
“Relative performance’’ denotes the percentage of the supervised upper-bound AP recovered by our method.  
Even for the challenging 50--80\,m range, the ego detector retains a strong fraction of the upper-bound performance, demonstrating that RSUs can provide high-quality pseudo-labels for distant objects.}
\label{tab:depth_range}
\begin{adjustbox}{width=.5\linewidth, center}
\begin{tabular}{lccc}
\toprule
Method & 0--30m & 30--50m & 50--80m \\
\midrule
Our pipeline & 85.0 & 81.4 & 76.0 \\
\textcolor{gray}{Upper bound} & \textcolor{gray}{97.1} & \textcolor{gray}{94.7} & \textcolor{gray}{86.5} \\ \midrule
\emph{Relative performance (\%)} & \emph{87.5} & \emph{86.0} & \emph{87.9} \\
\bottomrule
\end{tabular}
\end{adjustbox}
\end{table}

\begin{table*}[t]
\centering
\caption{
\textbf{Infrastructure-taught ego training on the \emph{rural} town.}
We analyze the impact of RSU supervision, communication noise, and pseudo-label refinement on ego detector performance.
``Refine'' denotes applying heuristic box refinement~\cite{luo2023drift} before ego training.
}
\label{tab:stage3_ego_town_7}
\begin{adjustbox}{width=.95\linewidth,center}
\begin{tabular}{l l c c c c c c}
\toprule
\multicolumn{2}{c}{Stage 1 (RSU training)} & \multicolumn{2}{c}{Stage 2 (Broadcasting)} & \multicolumn{4}{c}{Stage 3 (Ego testing)} \\
\cmidrule(lr){1-2} \cmidrule(lr){3-4} \cmidrule(lr){5-8}
Annotation source~~~~ & Method~~~~~~~~~~~~~~~~~~~~~~~~ & Comm.\ noise & Refine & ~Car~ & ~Ped~ & ~Cyc~ & ~\cellcolor{blue!10}Avg.~ \\
\midrule

\rowcolor{gray!10} \multicolumn{8}{l}{\emph{Ego detector: PointPillars~\cite{lang2019pointpillars}}}\\

12 RSUs (unsup.) & PP score + tracking & -- & -- 
& 81.9 & 76.6 & 58.7 & \cellcolor{blue!10}72.4 \\

12 RSUs (unsup.) & PP score + tracking & $\checkmark$ & -- 
& 81.7 & 23.7 & 30.7 & \cellcolor{blue!10}45.4 \\

\cmidrule{1-8}

12 RSUs (unsup.) & PP score + tracking & $\checkmark$ & $\checkmark$ 
& 83.0 & 63.1 & 31.3 & \cellcolor{blue!10}59.1 \\

\cmidrule{1-8}

\textcolor{gray}{12 RSUs (sup.)} & \textcolor{gray}{--} 
& \textcolor{gray}{--} & \textcolor{gray}{--} 
& \textcolor{gray}{91.0} & \textcolor{gray}{81.6} & \textcolor{gray}{76.5} & \cellcolor{blue!10}\textcolor{gray}{83.0} \\

\textcolor{gray}{12 RSUs (sup.)} & \textcolor{gray}{--} 
& \textcolor{gray}{$\checkmark$} & \textcolor{gray}{--} 
& \textcolor{gray}{92.4} & \textcolor{gray}{35.0} & \textcolor{gray}{46.5} & \cellcolor{blue!10}\textcolor{gray}{58.0} \\

\cmidrule{1-8}

\textcolor{gray}{Ego ground-truth} & \textcolor{gray}{--} 
& \textcolor{gray}{--} & -- 
& \textcolor{gray}{90.8} & \textcolor{gray}{84.9} & \textcolor{gray}{77.2} & \cellcolor{blue!10}\textcolor{gray}{84.3} \\

\midrule

\rowcolor{gray!10} \multicolumn{8}{l}{\emph{Ego detector: CenterPoint~\cite{yin2021centerpoint}}}\\

12 RSUs (unsup.) & PP score + tracking & -- & -- 
& 79.8 & 75.6 & 62.4 & \cellcolor{blue!10}72.6 \\

12 RSUs (unsup.) & PP score + tracking & $\checkmark$ & -- 
& 76.8 & 25.2 & 19.1 & \cellcolor{blue!10}40.4 \\

\cmidrule{1-8}

12 RSUs (unsup.) & PP score + tracking & $\checkmark$ & $\checkmark$ 
& 78.6 & 58.2 & 35.8 & \cellcolor{blue!10}57.5 \\

\cmidrule{1-8}

\textcolor{gray}{12 RSUs (sup.)} & \textcolor{gray}{--} 
& \textcolor{gray}{--} & \textcolor{gray}{--} 
& \textcolor{gray}{89.7} & \textcolor{gray}{70.9} & \textcolor{gray}{63.4} & \cellcolor{blue!10}\textcolor{gray}{74.7} \\

\textcolor{gray}{12 RSUs (sup.)} & \textcolor{gray}{--} 
& \textcolor{gray}{$\checkmark$} & \textcolor{gray}{--} 
& \textcolor{gray}{89.3} & \textcolor{gray}{25.6} & \textcolor{gray}{41.5} & \cellcolor{blue!10}\textcolor{gray}{52.1} \\

\cmidrule{1-8}

\textcolor{gray}{Ego ground-truth} & \textcolor{gray}{--} 
& \textcolor{gray}{--} & -- 
& \textcolor{gray}{86.5} & \textcolor{gray}{83.0} & \textcolor{gray}{69.2} & \cellcolor{blue!10}\textcolor{gray}{79.6} \\

\bottomrule
\end{tabular}
\end{adjustbox}
\end{table*}

\begin{figure}
    \centering
    \includegraphics[width=.8\linewidth]{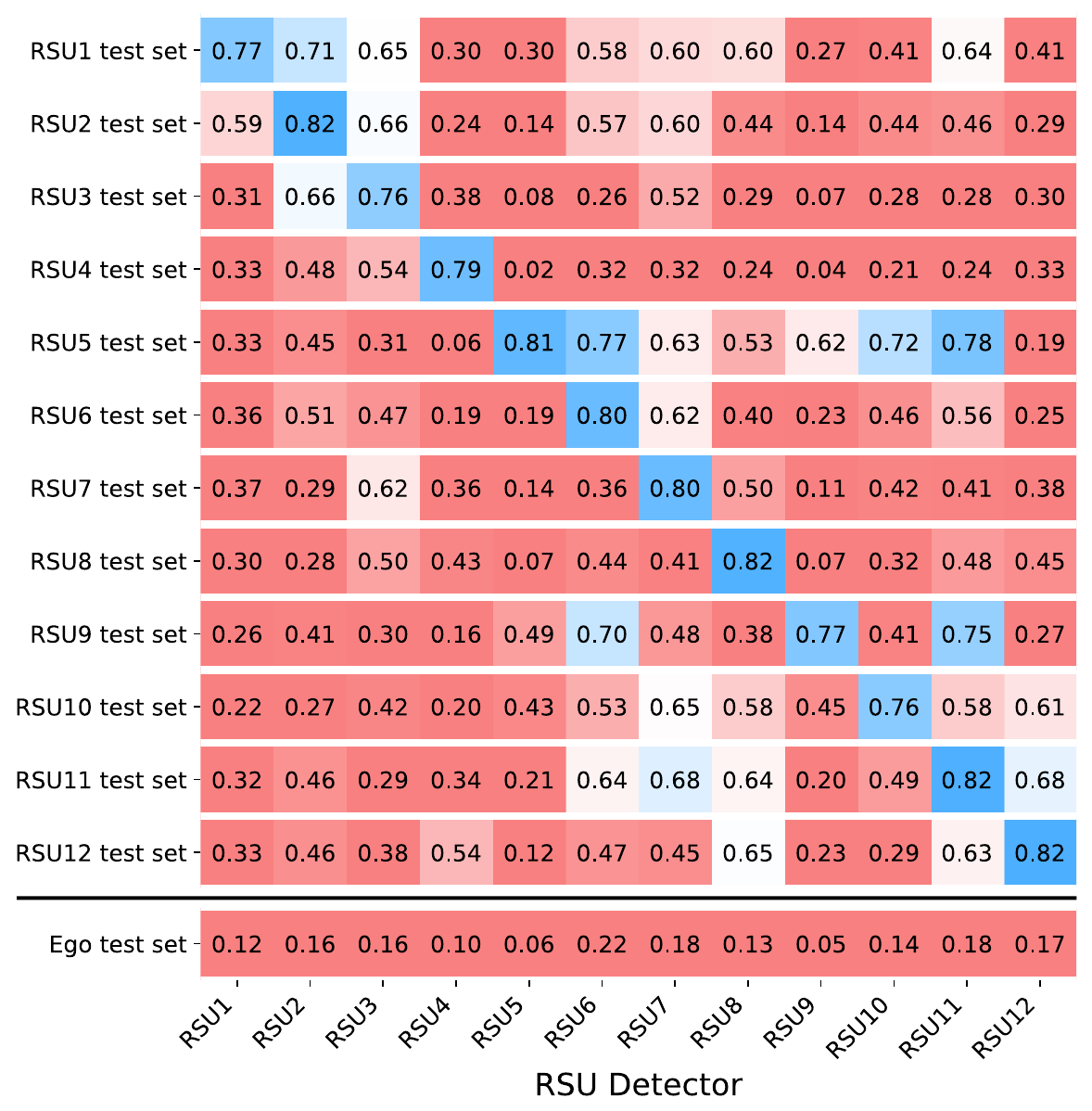}
    \caption{
\textbf{Cross-evaluation across RSUs in the rural town.}
Compared to the urban setting, RSU detectors exhibit slightly better cross-location generalization due to the simpler traffic semantics in rural environments.
Nevertheless, each RSU performs best within its own trained field of view and degrades when evaluated elsewhere.
The last row further shows that no single RSU detector transfers well to the ego viewpoint.
}
    \label{fig:rsu_cross_eval_rural}
\end{figure}

\mypara{Results on the rural town.}
As shown in~\cref{fig:rsu_cross_eval_rural}, unsupervised RSUs in the rural town exhibit slightly better cross-location generalization than those in the urban setting, likely due to the simpler scene structure. 
However, this generalization remains limited: each RSU performs best within its own fixed field of view and degrades noticeably when evaluated elsewhere. 
This observation highlights the inherently limited spatial coverage of individual infrastructure detectors and motivates collaborative infrastructure supervision.
Consistent with the observations from the urban scene, we also find that no RSU detector generalizes well to the ego viewpoint (last row of \cref{fig:rsu_cross_eval_rural}).

Quantitative results for the ego detector are reported in~\cref{tab:stage3_ego_town_7}. 
Due to computational constraints, we report only the core evaluations for the rural setting using two ego detectors~\cite{lang2019pointpillars,yin2021centerpoint}. 
Consistent with the findings from the urban setting in the main paper, the ego detector trained with infrastructure-generated pseudo-labels remains close to the supervised upper bound under ideal, noise-free communication for both detectors (\eg, 83.0\% AP for detecting vehicles compared to the upper bound of 90.8\% on PointPillars~\cite{lang2019pointpillars}). 
When communication noise is introduced, performance degrades, particularly for pedestrians and cyclists, reflecting the smaller number of pedestrian and cyclist instances available in the rural town for training. 
Nevertheless, simple heuristic-based box refinement~\cite{luo2023drift} effectively mitigates such communication noise (\eg, over 10 AP improvement on average and over 30 AP improvement for pedestrian detection for both detectors). 
Overall, these results further demonstrate the feasibility and effectiveness of infrastructure-taught ego training.

\mypara{Real-world feasibility study.}
Although existing real-world V2X datasets do not fully support our setup, we conduct a preliminary experiment to test whether our proposed mechanism holds. Specifically, we design a cross-intersection adaptation experiment on UrbanIng-V2X~\cite{sekaran2025urbaning}: the ego detector is pretrained on frames from all intersections except the target one, and is then adapted to the held-out intersection using our pipeline. With supervised RSU predictions as broadcast labels, dynamic object AP at IoU 0.3 improves from 23.5 to 29.3, indicating that the broadcast mechanism generalizes to real data. The unsupervised setting reaches 24.8; we attribute the remaining gap to MODEST~\cite{you2022modest}'s well-known difficulty with stationary objects like parked cars, which are abundant in the dataset.

\subsection{Qualitative Results}

\begin{figure}[t]
    \centering
    \includegraphics[width=\linewidth]{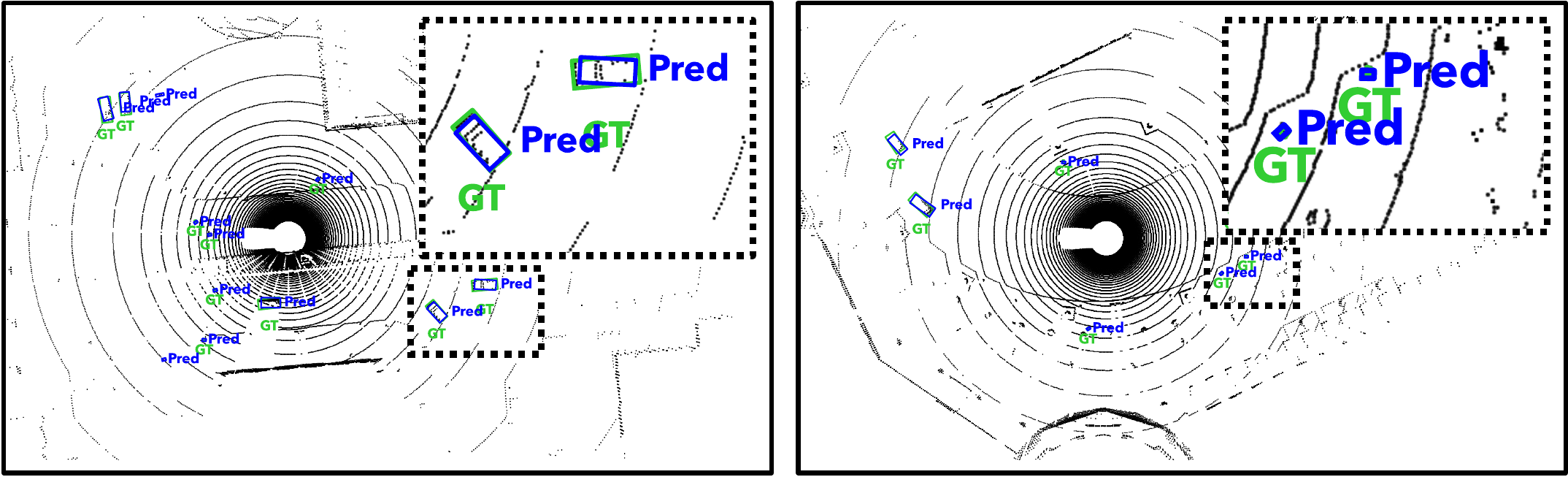}
   \caption{\textbf{Qualitative results of the ego detector with CenterPoint~\cite{yin2021centerpoint}.} Despite being trained without human labels, the ego detector generates accurate 3D bounding boxes, demonstrating the feasibility of our infrastructure-taught pipeline.}
    \label{fig:suppl_qualitative}
\end{figure}

\begin{figure}[t]
        \centering
        \includegraphics[width=.9\linewidth]{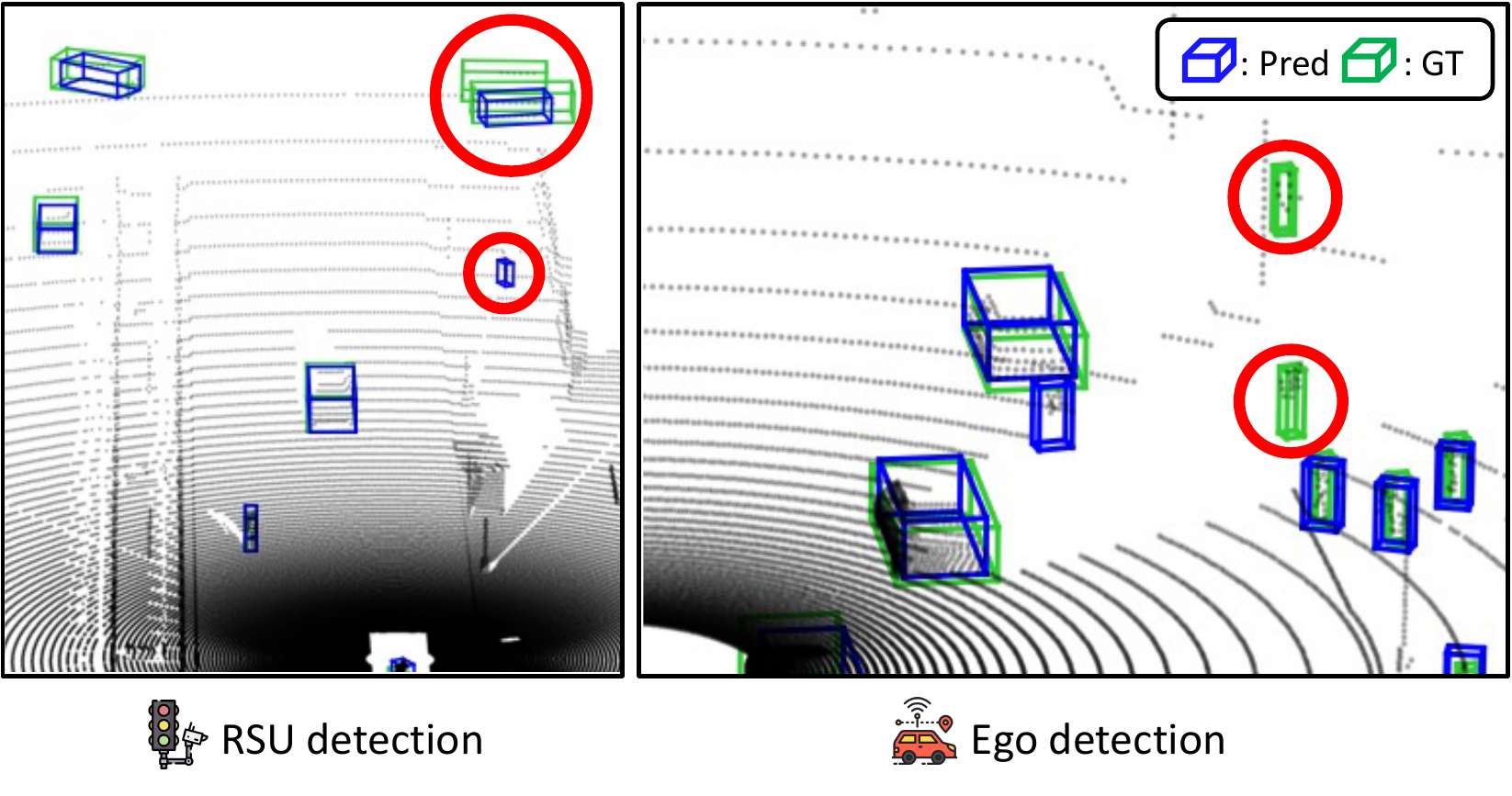}
        \vskip-10pt
        \caption{
\textbf{Failure cases.} While both RSU and ego detectors perform well in many cases, RSUs often miss distant objects and the ego detector struggles with small objects.
}
        \label{fig:suppl_quali_failure}
\end{figure}

\begin{figure*}
    \centering
    \includegraphics[width=\linewidth]{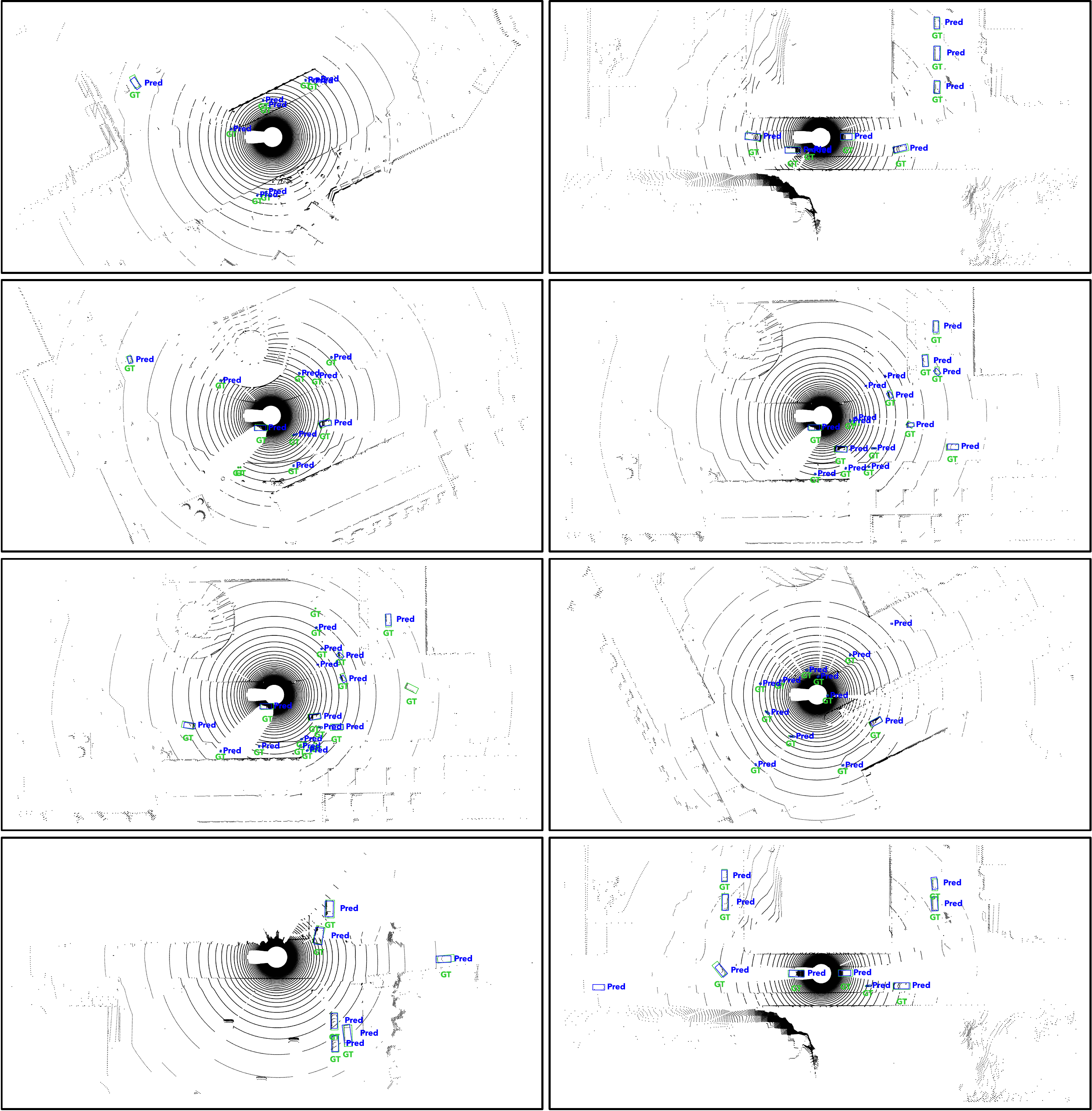}
    \caption{\textbf{Additional qualitative results of the ego detector. [1/2]}}
    \label{fig:suppl_quali_1}
\end{figure*}

\begin{figure*}
    \centering
    \includegraphics[width=\linewidth]{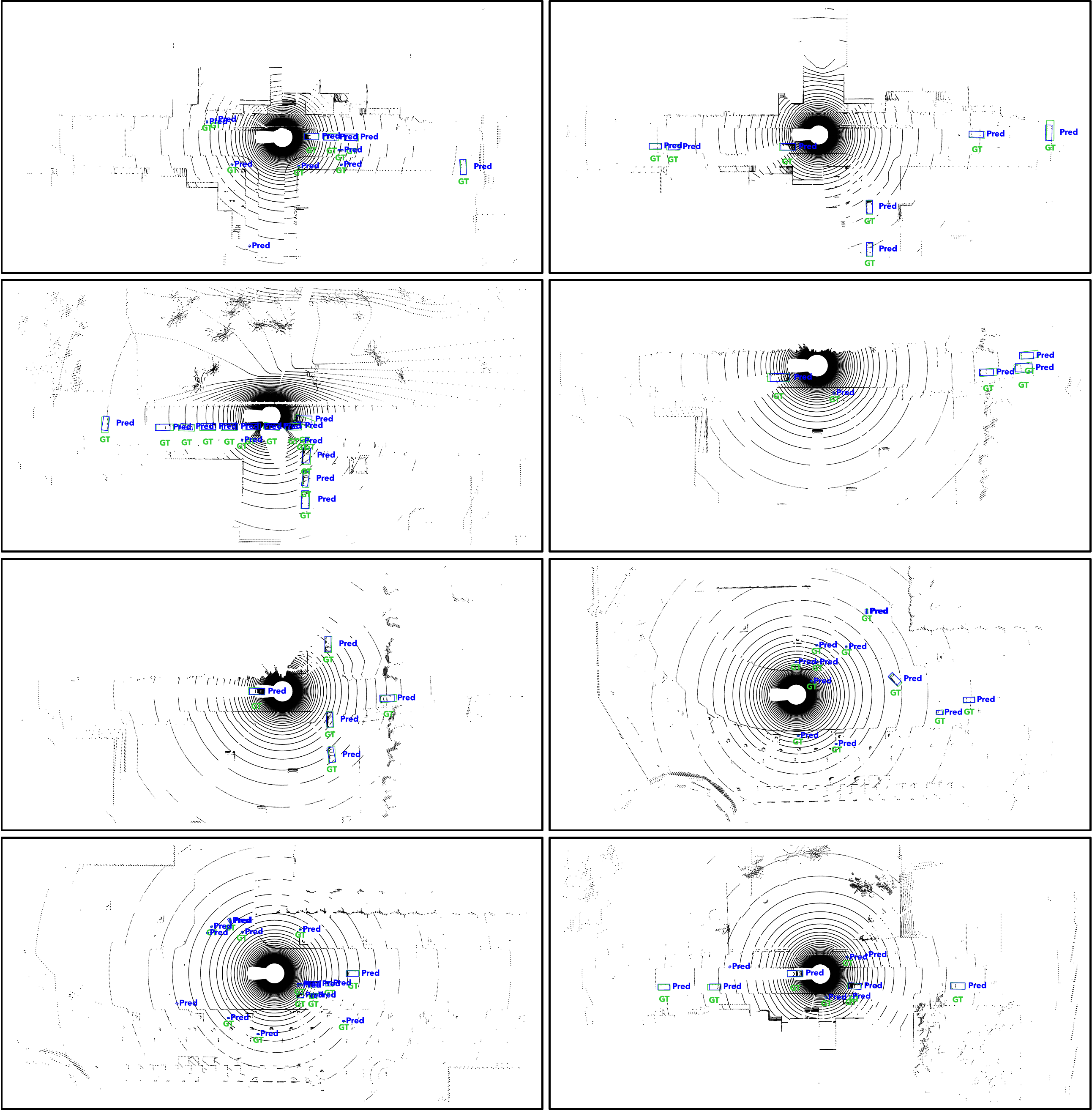}
    \caption{\textbf{Additional qualitative results of the ego detector. [2/2]}}
    \label{fig:suppl_quali_2}
\end{figure*}

Qualitative results in~\cref{fig:suppl_qualitative} show that the ego detector~\cite{yin2021centerpoint} produces accurate 3D bounding boxes.
Note that these results are obtained under simulated communication noise during RSU broadcasting, which better demonstrates the robustness of our proposed learning pipeline.
Additional uncurated examples are provided in~\cref{fig:suppl_quali_1} and~\cref{fig:suppl_quali_2}.
We further analyze failure cases in~\cref{fig:suppl_quali_failure}.
While both RSU and ego detectors perform well in many scenarios, RSUs may miss distant objects and the ego detector may struggle with small objects.
Overall, these qualitative examples further support the feasibility of infrastructure-taught, label-free 3D perception and highlight promising directions for future improvements.

% \clearpage

% \bibliography{main}

\end{document}